\documentclass[10pt,twocolumn,letterpaper]{article}

\usepackage{cvpr}              
\pdfoutput=1
\usepackage{graphicx}
\usepackage{amsmath}
\usepackage{amssymb}
\usepackage{booktabs}
\usepackage[pagebackref,breaklinks,colorlinks]{hyperref}

\usepackage[capitalize]{cleveref}
\crefname{section}{Sec.}{Secs.}
\Crefname{section}{Section}{Sections}
\Crefname{table}{Table}{Tables}
\crefname{table}{Tab.}{Tabs.}
\crefname{algocf}{Alg.}{Algs.}
\Crefname{algocf}{Alg.}{Algs.}
\usepackage{balance}
\usepackage{color}
\usepackage{xspace}

\newcommand{\method}{FlashEval\xspace}

\usepackage[utf8]{inputenc} 
\usepackage[T1]{fontenc}    
\usepackage{hyperref}       
\usepackage{url}            
\usepackage{booktabs}       
\usepackage{amsfonts}       
\usepackage{nicefrac}       
\usepackage{microtype}      
\usepackage[ruled,vlined]{algorithm2e}
\usepackage{algpseudocode}
\usepackage{enumitem}
\usepackage{multirow}
\usepackage[table]{xcolor}
\def\etal{{\em et al.~}}
\def\cvprPaperID{11437} 
\def\confName{CVPR}
\def\confYear{2024}

\begin{document}

\title{\method: Towards Fast and Accurate Evaluation of\\
Text-to-image Diffusion Generative Models}


\author{Lin Zhao$^{2*}$, Tianchen Zhao$^{12}\thanks{Equal contribution}$, Zinan Lin$^3$, Xuefei Ning$^{1\dagger}$, Guohao Dai$^{24}$, Huazhong Yang$^1$, Yu Wang$^{1}\thanks{Corresponding authors: Xuefei Ning and Yu Wang.}$\\
$^1$Tsinghua University, $^2$Infinigence-AI,  $^3$Microsoft Research, $^4$Shanghai Jiao Tong University\\
{\tt\small \{lllzz0309zz,suozhang1998,linzinan1995,foxdoraame\}@gmail.com \quad daiguohao@sjtu.edu.cn} \\ \tt\small\{yanghz, yu-wang\}@mail.tsinghua.edu.cn \\
}
\maketitle
\begin{abstract}
In recent years, there has been significant progress in the development of text-to-image generative models. Evaluating the quality of the generative models is one essential step in the development process. 
Unfortunately, the evaluation process could consume a significant amount of computational resources, making the required periodic evaluation of model performance (e.g., monitoring training progress) impractical.
Therefore, we seek to improve the evaluation efficiency by \textbf{selecting the representative subset of the text-image dataset}. 
We systematically investigate the design choices, including the selection criteria (textural features or image-based metrics) and the selection granularity (prompt-level or set-level). 
We find that the insights from prior work on subset selection for \textbf{training} data do not generalize to this problem, and we propose \method{}, an iterative search algorithm tailored to \textbf{evaluation} data selection. 
We demonstrate the effectiveness of \method{} on ranking diffusion models with various configurations, including architectures, quantization levels, and sampler schedules on COCO and DiffusionDB datasets.
Our searched 50-item subset could achieve comparable evaluation quality to the randomly sampled 500-item subset for COCO annotations on unseen models, achieving a 10x evaluation speedup.
We release the condensed subset of these commonly used datasets to help facilitate diffusion algorithm design and evaluation, and open-source \method{} as a tool for condensing future datasets, accessible at \url{https://github.com/thu-nics/FlashEval}.

\end{abstract}

\begin{figure}[t]
    \centering
    \includegraphics[width=1.0\linewidth]{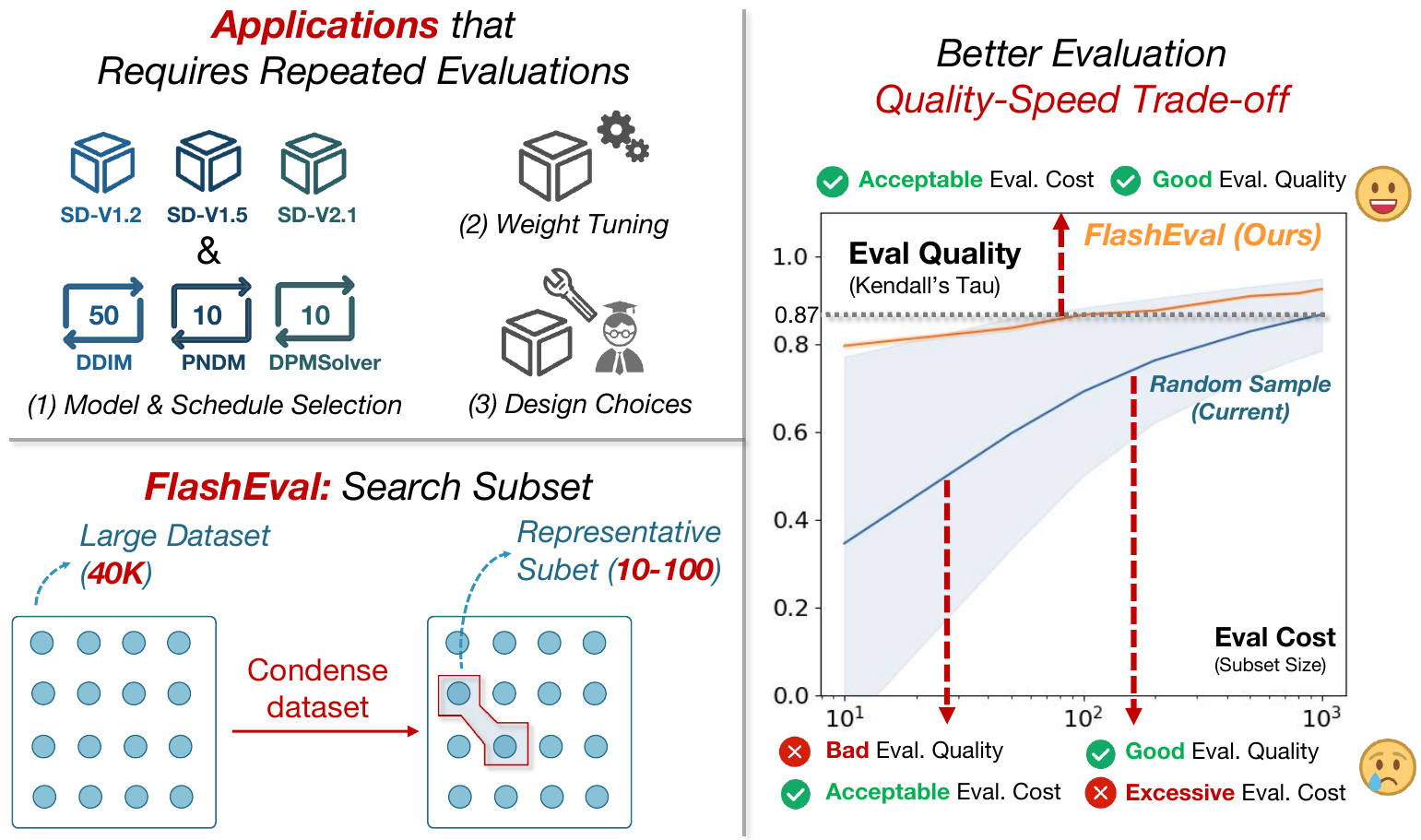}
    \caption{\textbf{The Motivation and Effectiveness of \method .} Left: Applications and the search method. Right: The excessive evaluation cost and how \method improves the evaluation speed-quality trade-off (evaluation quality is represented by the ranking correlation of a variety of diffusion models w.r.t subset size on COCO).}
    \label{fig:teaser}
    \vspace{-0.2cm}
\end{figure}

\section{Introduction}
\label{sec:introduction}


Diffusion models could generate high-fidelity images based on textual descriptions (referred to as prompts) and achieve substantial research interest \cite{stable-difffusion, GLIDE, ImageReward, Imagen}.
Researchers have designed a range of techniques aimed at enhancing these models from various angles. 
Some algorithm design applications often necessitate a large number of iterative evaluations, such as:
\textbf{(1) Model $\&$ Schedule Selection:} The quality of image generation exhibits varying trends over different timesteps for different models and solvers. Consequently, evaluations and comparisons are needed to strike a balance between generation quality and computational cost.
\textbf{(2) Weight Tuning:} In the process of training and fine-tuning the model, evaluating the performance of current parameters is required for refining the training strategy.
\textbf{(3) Design Choices:} The iterative testing of multiple design options, such as layer-wise pruning rates, layer-wise mixed precision configurations, and other hyperparameter choices. 

To evaluate the diffusion model, prior research~\cite{OMS-DPM,valid_loss_1,valid_loss_2} points out that the denoising loss does not necessarily reflect the generation quality. 
Therefore, existing evaluation schemes estimate the image quality (e.g., FID~\cite{FID}, IS~\cite{IS}, ImageReward~\cite{ImageReward}) over text annotations of existing text-image datasets~\cite{COCO,DiffusionDB} or manually-designed textual dataset~\cite{Pick-a-pic} (we refer them as ``test prompts''). 
However, as test prompts have a relatively large size (e.g., COCO~\cite{COCO} 47K, PicScore~\cite{Pick-a-pic} 15K), the evaluation process has a significant computational burden.\footnote{ 
For example, evaluating stable diffusion V1.5 on the whole COCO dataset requires 60 GPU hours (RTX 3090).} 
Iterative evaluations required in the model design phase are thus impractical, and full set evaluation is often only adopted in the final evaluation (e.g., in DALL-E~\cite{DALL-E} and Stable-Diffusion~\cite{stable-difffusion}).
To circumvent this challenge, the  common practice is to 
\textbf{randomly sampling a smaller subset} (e.g., 1K prompts in \cite{X-IQE, OMS-DPM}) for a proxy evaluation.
However, we find that such random sub-sampling exhibits a poor accuracy-efficiency trade-off. 
In \cref{fig:teaser}, we use this approach to evaluate a model zoo with various schedules. 
Even with $1000$ samples, this approach only achieves a rank correlation of $0.87$ (value of 1 indicates perfect correlation).
More aggressive sub-sampling (for more speed-up) would result in an acute drop in ranking correlation and a large variation of results.

In this paper, \textbf{we aim to improve the accuracy-efficiency trade-off of diffusion evaluation by seeking ``representative subsets'' of the test prompts.} 
Firstly, we systematically explore the potential characteristics in representative sets. 
Then, we design baseline search methods to validate these characteristics and delve into the reasons for their failure when dealing with small item sizes.
We conclude that the key issue leading to suboptimal search results is the inadequate exploration of the vast search space.
Finally, inspired by the evolutionary algorithm, we design an improved iterative search algorithm on both the set and prompt levels.
The proposed \method search method could effectively mitigate variance and improve evaluation quality under small item sizes. 
Our acquired 50-item subset demonstrates a ranking correlation comparable to that of a 200-item baseline-acquired subset, and a randomly sampled 500-item subset.
We will release the condensed subsets of different sizes for the commonly used datasets to assist researchers in selecting the appropriate prompt quantity for evaluation. Additionally, 
\method{} could be used to search condensed evaluation subsets for new datasets.

In summary, our contributions can be listed as follows:

\begin{itemize}
\item We identify the need to improve the accuracy-efficiency trade-off of diffusion model evaluation and introduce the idea of \textbf{identifying a representative subset} to speed up evaluation. 

\item We systematically investigate the potential properties of representative sets and propose an improved search method to effectively acquire representative subsets even with small item size.

\item We aim not to replace but to provide an accurate proxy of the existing evaluation scheme. We hope that \method could assist researchers in selecting the proper prompt quantity, and help accelerate and facilitate the broader diffusion algorithm design.
\end{itemize}

\section{Related Work}
\label{sec:rw}
\noindent\textbf{Metrics for Text-to-Image Generation.}
The currently widely used metrics 
are often on 
one or more of the following aspects: image fidelity, image-text alignment, and image aesthetic.
To assess fidelity, FID \cite{FID} and IS \cite{IS} are commonly used metrics that measure the feature distance between generated images and real images.
CLIPScore \cite{CLIPScore} is widely used to measure image-text alignment. It calculates the similarity of features extracted from the image and text domains by dedicated encoders, respectively.
As for aesthetic, Schuhmann \etal~\cite{Aesthetic} train a score predictor using a dataset dedicated to aesthetics to facilitate evaluations.

To better align metric scores with human preference, recent methods perform manual scoring for generated images and train a predictor capable of evaluating models.
PickScore~\cite{Pick-a-pic} and HPS~\cite{HPS} generate two or more images for a given prompt, enabling users to choose the best one.
ImageReward~\cite{ImageReward} involves users comprehensively considering both alignment and fidelity to assign explicit scores to each image.
We 
apply metrics for all the aspects mentioned above, validating that FlashEval can be efficiently utilized to assess various aspects of models.

\noindent\textbf{Diffusion Model Evaluation Datasets.}
To comprehensively evaluate Text-to-Image (T2I) models, there are currently various datasets available. Among them, the COCO validation set \cite{COCO} is one of the most widely used datasets~\cite{GLIDE,guo2022assessing,guo2023zero,X-IQE}.
In addition, to enhance benchmark precision, Saharia \etal~\cite{drawbench} conduct the first attempt to consider multiple evaluation aspects and propose DrawBench, which contains $200$ prompts specifically designed for evaluating T2I tasks.
Cho \etal~\cite{Dall-eval} introduce PAINTSKILLS, an evaluative dataset crafted to assess three essential visual reasoning skills: object recognition, object counting, and spatial relationships.
%
%
Kirstain \etal~\cite{Pick-a-pic} create a web app to build a large, open dataset of T2I prompts and real users’ preferences over generated images.
Chen \etal~\cite{X-IQE} utilize GPT to provide scores and more detailed evaluative descriptions.

Recently, Bakr \etal~\cite{HRS} categorize different metrics and design distinct datasets for each category, aiding users in evaluating specific metrics more effectively.
Similarly, Lee \etal~\cite{holistic} craft different scenarios and propose various datasets for each scenario. 
For any aforementioned datasets that are too large to evaluate efficiently, \method{} can 
be applied to enhance
the efficiency of evaluating models. 

\noindent\textbf{Training/Test Set Selection.} 
There is a category of methods called "coreset selection" that accelerates the model training process by choosing subset from the training data, which aim to retain challenging examples in the subset \cite{data-effiency,Squeeze,bilevel}.
Depending on the implementation, they can be categorized into two types: supervised learing \cite{forgetting,EL2N,Memorization} and unsupervised learning \cite{K-center,Kmeans,unsupervised}. 
For supervised learning methods, it is common to assess each example during the training process to decide whether it should be included in the subset.
Toneva \etal~\cite{forgetting} track how often each example is forgotten during training. They assume that those forgotten less often have a smaller impact on the final results.
Paul \etal~\cite{EL2N} believe examples with high training loss should be retained, which indicate difficulty in learning.
They train for only a few epochs for filtering, aiming to reduce the cost of subset selection.
Unsupervised methods often select subsets based on clustering results.
Ozan \etal~\cite{Kmeans} argue that examples farther from the centroid present greater learning challenges and should be retained.
Sorscher \etal~\cite{K-center} use the K-center to ﬁnd a diverse cover for active learning.

There are also some approaches that intend to reduce testing cost and do unbiased evaluation with unlabeled test dataset.
Kossen \etal~\cite{active1} actively select examples to be labeled by learning a surrogate model that directly predict losses at all data.
Deng~\cite{active2} obtain the transformed images and labels from training data to train a regression model for predicting scores of unlabeled data. 
Unlike them, our method wants to find the subset that are representative of the entire validation dataset.
Recent work by \cite{Anchor} attempts to obtain the subset for evaluating language models.
To best of our knowledge, we are the first to identify representative subset to accelerate evaluation for T2I models.





\section{Problem Formulation}
\label{sec:problem_formulation}

In this section, we provide a brief overview of the problem formulation for \textbf{``identifying a representative subset''} that has a good accuracy-efficiency trade-off.
We choose the application scenario of ranking a variety of models (which is typically the case in the examples in \cref{sec:introduction}), and use the \textbf{ranking correlation} for measuring the evaluation quality. 
More specifically, we consider a set of models $N_m$ models: $\mathcal{M}=\{M_i|i\in\{1,\ldots,N_m\}\}$
(with diverse model architectures, parameters, solvers, and/or step sizes). 
Given a dataset $\mathcal{P} = \{p_i | i \in \{1,...,N\}\}$ of size $N$, we aim to find a representative subset $\hat{\mathcal{P}} = \{ p_i | i \in \mathcal{I} \}$ of item size $N'$ with index list $\mathcal{I} = \{i_1, ..., i_{N'}\}$ that maximizes the evaluation quality. 
For single-image evaluation metrics (e.g., CLIPScore, HPS, ImageReward), the metric score of prompt $p_i$ for model $M_j$ is described by $\mathcal{S}_{ij}$. We use the averaged metric scores on the entire textual dataset as the ``ground-truth performance'' of certain model $\hat{\mathcal{S}_j} = \sum_{i=1}^{N} \frac{1}{N} \mathcal{S}_{ij}$. For multiple-images metrics (e.g., FID), we use the full set evaluation as ground-truth. 
We adopt Kendall's Tau (KD) \cite{kd} to measure the ranking correlation, which is computed as follows.
Let's denote the ground-truth metric and the approximated metric (computed using $\hat{\mathcal{P}}$) as $x_1,...,x_{N_m}$ and $y_1,...,y_{N_m}$. For each pair of models $M_i$ and $M_j$, we denote them as a \emph{concordant} pair if $x_i<x_j$ and $y_i<y_j$, or $x_i>x_j$ and $y_i>y_j$ (i.e., they have a consistent ranking). Similarly, we denote them as a \emph{discordant} pair if $x_i<x_j$ and $y_i>y_j$, or $x_i>x_j$ and $y_i<y_j$. 
Among all pairs of models, we count the number of concordant and discordant pairs as $N_c$ and $N_d$, respectively. Furthermore, we define $n_1$ as the number of pairs that have a tie only in the ground-truth metric (i.e., $x_i=x_j$ and $y_i\not=y_j$), and $n_2$ as the number of pairs that have a tie only in the approximated metric. Here, metric equivalence is deemed if the difference between values is smaller than a certain threshold, which is determined by the 3 times standard deviation across the scores of 50 randomly generated images from the same model. 
Given the values of $N_c,N_d, n_1$, and $n_2$, KD is computed as 
\begin{equation}
\begin{split}
        \frac{N_c - N_d}{\sqrt{(N_c + N_d+n_1)(N_c + N_d+n_2)}},
\end{split}
\label{equ:kd}
\end{equation}


\section{Methods}
\label{sec:method}
In T2I tasks, we can identify subsets from both the textual feature space and image metric space.
In this section, we systematically investigate the potential properties of the representative subsets in the two spaces. 
Then, we design baseline methods to use these
properties and analyze the reasons for their failure under smaller item sizes. 
Finally, we present an improved search algorithm that addresses such issues in Sec.~\ref{sec:methods_ours}.

\subsection{Baseline Approaches for Subset Selection}
\label{sec:method_baseline}
%
A straightforward way of identifying the representative subset is by seeking samples that cover the entire dataset in terms of categories (\textbf{Baseline 1}) or distribution (\textbf{Baseline 2}) in the \textbf{textual feature space} following previous methods~\cite{K-center,data-quan}.
However, abundant experiments in the appendix show that relying solely on textual features is not adequate for identifying the representative subset. 

Therefore, we turn to examine the characteristics of the direct metric scores of generated images (\textbf{image-based metrics}). 
We propose to directly optimize the ranking correlation, and design search-based algorithms on the set and prompt level and validate their effectiveness on relatively medium-sized subsets. However, for smaller subset sizes like 10-50, their identified subsets exhibit significant variance and may yield poor performance. We further analyze the reasons for their failure. 

\textbf{Baseline 3: KD-based random search algorithm.}
To identify a representative set, we introduce a straightforward KD-based random search algorithm.
There are two units for searching—set-wise search and prompt-wise search. For \textbf{set-wise search}, the fundamental search unit is ``set''. The process involves sampling a large number of (e.g., 1 million) \textbf{subsets} randomly from the entire dataset $\mathcal{P}$, and then identifying the subset $\hat{\mathcal{P}}$ that yields best KD. 
For \textbf{prompt-wise search}, the fundamental search unit is ``prompt''. We find the top $K$ \textbf{prompts} with the highest KD to identify the representative subset with $K$ prompts. 
We randomly split the model zoo to be ranked as training/testing, and adopt the above mentioned search on the training models. 

\begin{table}[t]
\small
\centering
\vspace{-0.5cm}
\resizebox{0.8\linewidth}{!}{\begin{tabular}{c|c|cccc}
\toprule
\multicolumn{1}{c|}{\multirow{2}{*}{Methods}} & \multirow{2}{*}{Type} & \multicolumn{2}{c}{$N'=10$} & \multicolumn{2}{c}{$N'=100$} \\ \cmidrule(l){3-6} 
\multicolumn{1}{c|}{} &  & Train & Test & Train & Test \\ 
\midrule
\midrule
RS & - & 0.355 & 0.346 & 0.706 & 0.692 \\
\midrule
B3-set & \multirow{3}{*}{\begin{tabular}[c]{@{}c@{}}Image\\ Metrics\end{tabular}} & 0.827 & 0.603 & 0.925 & 0.800 \\
B3-prompt &  & 0.822 & 0.610 & 0.914 & 0.764 \\
Ours &  & \textbf{0.934} & \textbf{0.796} & \textbf{0.966} & \textbf{0.867} \\ 
\midrule
\end{tabular}}
\caption{\textbf{Comparisons of the Kendall's Tau of CLIP-Score for subsets acquired by \method and baseline methods.} The ``RS'' represents sampling $N'$ prompts randomly from dataset.} 
\label{tab:baselines}
\vspace{-0.3cm}
\end{table}

\begin{figure}[t]
    \centering
    \vspace{0cm}
    \includegraphics[width=0.8\linewidth]{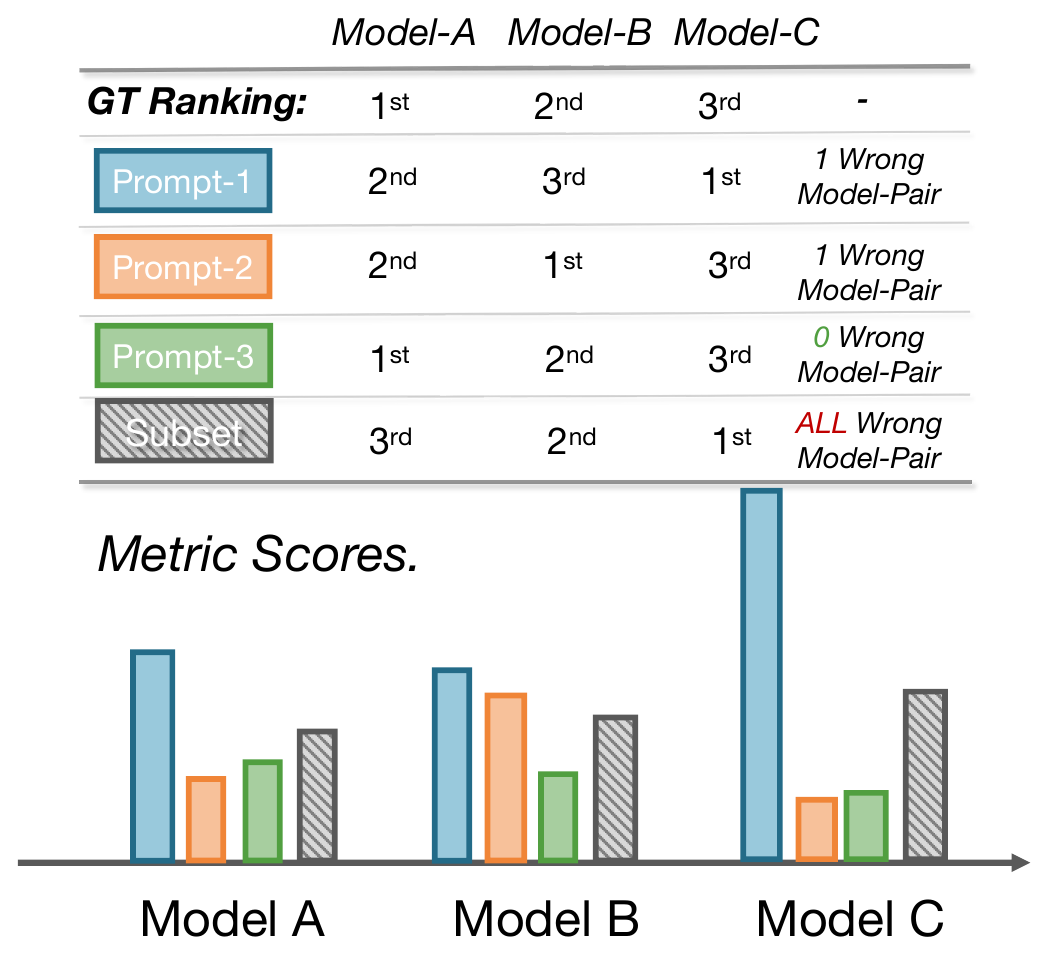}
    \caption{\textbf{Illustration of the reasons for baseline-3's failing under small item sizes $N'$.} When combining multiple prompts with standalone high KD, the set-wise KD is lower. }
    \label{fig:bad_cases}
    \vspace{-0.5cm}
\end{figure}

\textbf{Results for KD-based search.}
As could be seen from ~\cref{tab:baselines}, the results of prompt-wise search (brief for \textbf{``B3-prompt''}) and set-wise search (brief for \textbf{``B3-set''}) achieve better results than RS.
When $N'=100$, the searched subset achieves a KD of 0.75-0.8 on test set, whereas when $N'=10$, the KD is worse ($\approx$0.6).

\textbf{Analysis of why B3-prompt failed.}
For ``B3-prompt'', we conduct detailed analysis for searched prompt and composed set. 
When combining multiple prompts with standalone high KD, the set-wise ranking KD notably decreases. 
As shown from the toy examples in 
\cref{fig:bad_cases}: the prompt-1,2,3 have high standalone KD (rank one or none of the models wrong), but the combined subset have low KD (rank all model wrong). 
It reveals that searching the top prompts and combining them may not produce the optimal subset.

\textbf{Analysis of why B3-set failed.}
Based on the analysis above, the set-wise search might be a better approach because it takes the interference between prompts into consideration.
However, the improvement of set-wise search over prompt-wise search is marginal
even on the training set as shown in \cref{tab:baselines}.
We attribute such a phenomenon to the insufficient exploration of the search space. Take $N'=10$ as example, the complete search space of constructing 10 prompts from $40{,}504$ text annotations (COCO dataset), the search space size is $C_{40504}^{10}$, which is impractical to traverse through. Even we construct numerous subsets (i.e. 1M), it is still restricted compared with the enormous search space size. Therefore, an improved search algorithm with enhanced sample efficiency is required to generate representative subsets. 



\begin{figure}[t]
    \centering
    \vspace{-0.6cm}
    \includegraphics[width=0.9\linewidth]{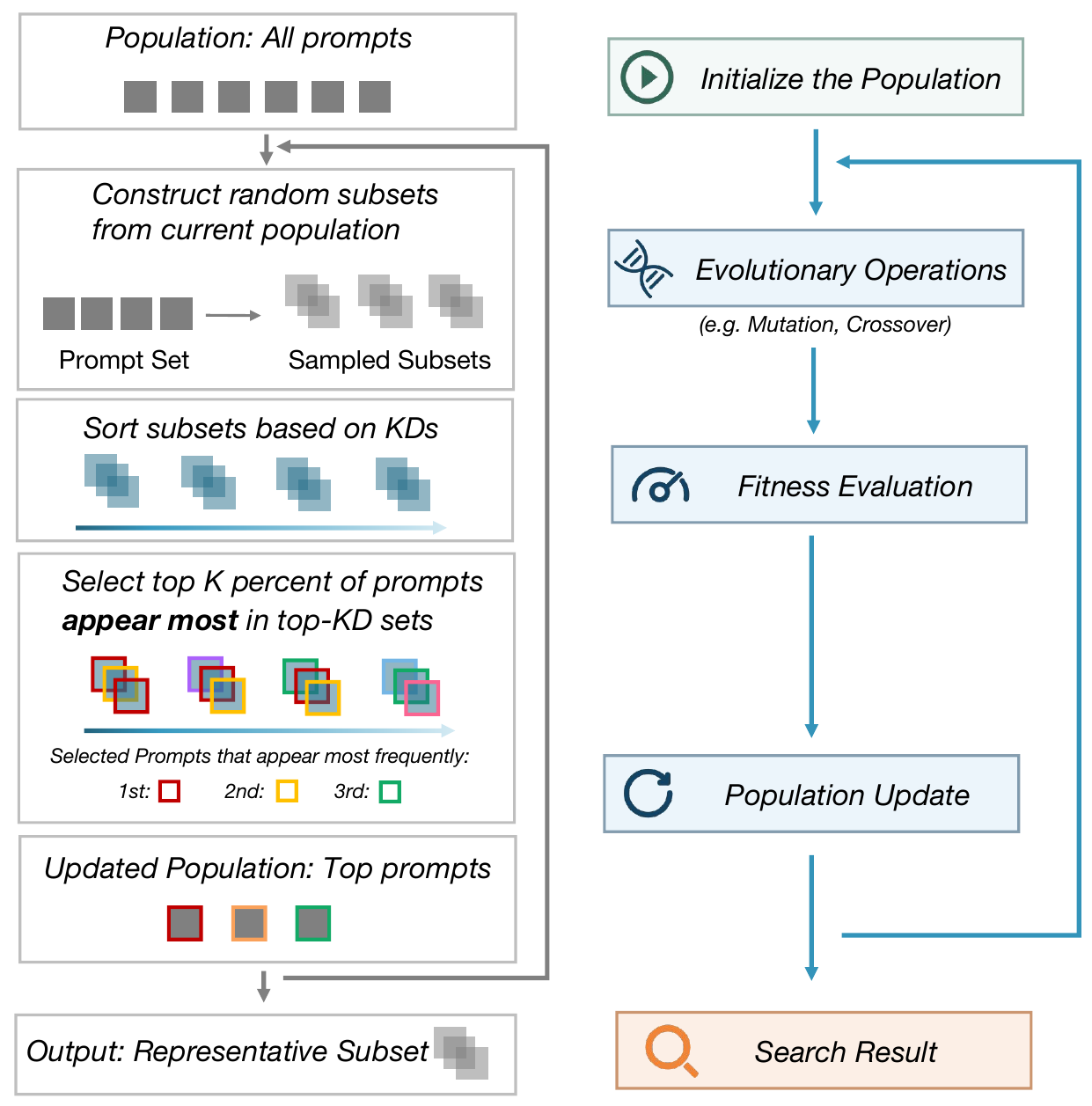}
    \caption{\textbf{Illustration of \method search method.} Inspired by the evolutionary algorithm, we design an iterative search algorithm on both the set- and prompt-level.}
    \label{fig:evo}
    \vspace{-0.5cm}
\end{figure}

\subsection{\method Search Method}
\label{sec:methods_ours}

As discussed above, the prompt-wise search is efficient but results in suboptimal subsets because it doesn't consider interactions between prompts. The set-wise search 
considers the interactions
but suffers from insufficient exploration of enormous search space. The key problem is to improve the sample efficiency of set-wise searching. In light of the prior that well-performing prompts are more likely to construct well-performing sets, we propose to utilize the prompt-level information to provide guidance for the set-level search. Inspired by the evolutionary algorithms~\cite{evolutionary,evolutionary_algorithm_wikipedia}, we design an iterative search method on both the prompt and set level. We provide an overall introduction to the workflow of our method in \cref{sec:method_iter}. Then, we elaborate on our frequency-based filtering method in \cref{sec:method_freq}.
The detailed description of \method search algorithm flow is listed in Algo.~\ref{algo:search}.


\begin{algorithm}[t]
    \caption{\method Search Algorithm}
    \resizebox{0.9\linewidth}{!}{ 
    \begin{minipage}{0.99\linewidth}
    \textbf{Input:} Whole textual dataset $\mathcal{P} = \{p_i | i=1,...,N\}$, The train split of the model zoo $\mathcal{M}_{train}$, the number of candidate subsets to rank $N_{set}$, the number of iterations $N_{iter}$, keep ratio of prompts $K_{p}$, keep ratio of sets $K_{s}$;
  
    \textbf{Ouput:} Representative subsets $ \hat{\mathcal{P}}$ of size $N'$;

    Get the model ranking $R_{gt}$ on $\mathcal{P}$;

    \tcp{Initialize the population}
    
    Initialize $\mathcal{P'} \gets \mathcal{P}$;
    
    \For{$i \gets 1$ to $N_{iter}$}{
        \For{$j \gets 1$ to $N_{set}$}{

            \tcp{Evolutionary Operations}
            
            Construct random subset $\mathcal{P}_{j} \sim \mathcal{P'}$ of size $N'$, get its model ranking $R_j$ on $\mathcal{M}_{train}$\;
            Calculate $\mbox{KD}_j$ values between $R_j$ and $R_{gt}$\;
        }

        \tcp{Fitness Evaluation}
        
        $S_{set}\gets$ subsets among $\{\mathcal{P}_{j}\}_{j=1}^{N_{set}}$ with top $K_s$ percent of KD\;

        \tcp{Population update}

        $S_{prompt}\gets$ top $K_p$ percent of prompts  that appear most frequently in $S_{set}$; 
            
        Update candidate prompt set $\mathcal{P'} \gets S_{prompt}$;
  }  
  Construct subsets $\{\mathcal{P}_j\}_{j=1}^{N_{set}}$ of size $N'$ from $\mathcal{P'}$\;
  
  \Return $\hat{\mathcal{P}}$ with best KD in $\{\mathcal{P}_j\}_{j=1}^{N_{set}}$
  \end{minipage}
  }
  \label{algo:search}
\end{algorithm}

\begin{table*}
\vspace{-0.5cm}
\setlength{\abovecaptionskip}{0.1cm}
\begin{center}
\raisebox{3pt}{
\resizebox{0.9\linewidth}{!}{
\begin{tabular}{c c c c c c c c}
\hline
\toprule[1pt]
\multirow{2}{*}{models} & {item size} & \multicolumn{3}{c}{$N'$=50}  & \multicolumn{3}{c}{$N'$=500}  \\
\cmidrule(lr){2-2} \cmidrule(lr){3-5} \cmidrule(lr){6-8}
 & methods \textbackslash sub-tasks & random & model variants &  schedulers & random & model variants & schedulers\\
\midrule\midrule
\multirow{5}{*}{Train} & {RS} & 0.607$\pm$0.000 & 0.594$\pm$0.000 & 0.632$\pm$0.000 &  0.857$\pm$0.000 & 0.858$\pm$0.000 & 0.857$\pm$0.000 \\ 
 & {B3-prompt} & 0.900 & 0.909 & 0.862 & 0.895 &0.917 & 0.872 \\
 & {B3-set} & 0.895$\pm$0.002 & 0.912$\pm$0.002 & 0.894$\pm$0.002 & 0.971$\pm$0.002 & 0.970$\pm$0.001 & 0.966$\pm$0.002 \\  
 & {Ours} & \textbf{0.956$\pm$0.004} & \textbf{0.969$\pm$0.004} & \textbf{0.960$\pm$0.004}  & \textbf{0.984$\pm$0.003} & \textbf{0.986$\pm$0.003} & \textbf{0.978$\pm$0.003} \\ 
\midrule 
\multirow{5}{*}{Test} & {RS} & 0.597$\pm$0.000 & 0.588$\pm$0.000 & 0.560$\pm$0.000 & 0.829$\pm$0.000 & 0.826$\pm$0.000 & 0.827$\pm$0.000  \\
 & {B3-prompt} & 0.729 & 0.784 & 0.810 & 0.805 & 0.822 & 0.851 \\
 & {B3-set} & 0.750$\pm$0.014 & 0.680$\pm$0.021 & 0.721$\pm$0.013  & 0.875$\pm$0.007 & 0.836$\pm$0.008 & 0.863$\pm$0.008 \\ 
 & {Ours} & \textbf{0.851$\pm$0.004} & \textbf{0.800$\pm$0.008} & \textbf{0.850$\pm$0.005}  & \textbf{0.906$\pm$0.003} & \textbf{0.899$\pm$0.004} & \textbf{0.909$\pm$0.003} \\ 
\bottomrule[1pt]
\end{tabular}
}
}
\end{center}
\caption{\textbf{Comparisons of the Kendall's Tau of CLIP-Score on COCO dataset (the error bars are standard errors)}. B3-prompt only requires finding prompts once, hence there's no need to calculate standard errors. ``Random'', ``model variants'', ``schedulers'' are the three sub-tasks mentioned in \ref{sec:sub-task}. The ``RS'' represents sampling $N'$ prompts randomly from dataset.}
\label{tab:COCO}
\end{table*}

\begin{table*}
\begin{center}
\resizebox{0.9\linewidth}{!}{
\begin{tabular}{c c c c c c c c}
\hline
\toprule[1pt]
\multirow{2}{*}{models} & {item size} & \multicolumn{3}{c}{$N'$=5}  & \multicolumn{3}{c}{$N'$=100}  \\
\cmidrule(lr){2-2} \cmidrule(lr){3-5} \cmidrule(lr){6-8}
 & methods \textbackslash sub-tasks & random & model variants &  schedulers & random & model variants & schedulers \\
\midrule\midrule
\multirow{5}{*}{Train} & {RS} & 0.395$\pm$0.000 & 0.307$\pm$0.000  & 0.413$\pm$0.000 & 0.730$\pm$0.000 & 0.721$\pm$0.000 & 0.755$\pm$0.000 \\
 & {B3-prompt} & 0.819 & 0.725 & 0.820 & 0.879 & 0.843 & 0.920 \\ 
 & {B3-set} & 0.839$\pm$0.003 & 0.827$\pm$0.003 & 0.833$\pm$0.002 & 0.935$\pm$0.002 & 0.924$\pm$0.001 & 0.933$\pm$0.002 \\ 
 & {Ours} & \textbf{0.908$\pm$0.004} & \textbf{0.891$\pm$0.004} & \textbf{0.897$\pm$0.006}  & \textbf{0.966$\pm$0.002} & \textbf{0.968$\pm$0.002} & \textbf{0.967$\pm$0.003} \\ 
\midrule 
\multirow{5}{*}{Test} & {RS} & 0.350$\pm$0.000 & 0.423$\pm$0.000 & 0.323$\pm$0.000 & 0.763$\pm$0.000 & 0.681$\pm$0.000 & 0.738$\pm$0.000 \\   
 & {B3-prompt} & 0.520 & 0.266 & 0.657 & 0.752 & 0.683 & 0.806 \\
  & {B3-set} & 0.643$\pm$0.031 & 0.516$\pm$0.033 & 0.601$\pm$0.018 & 0.826$\pm$0.010 & 0.792$\pm$0.018 & 0.836$\pm$0.014 \\ 
 & {Ours} & \textbf{0.789$\pm$0.009} & \textbf{0.691$\pm$0.011} & \textbf{0.748$\pm$0.014}  & \textbf{0.851$\pm$0.007} & \textbf{0.861$\pm$0.007} & \textbf{0.873$\pm$0.007} \\
\bottomrule[1pt]
\end{tabular}
}
\end{center}
\vspace{-0.3cm}
\caption{\textbf{Comparisons of the Kendall's Tau of CLIP-Score on DiffusionDB dataset (the error bars are standard errors).}}
\label{tab:diffusionDB}
\end{table*}

\subsubsection{Set- and Prompt-level Iterative Filtering}
\label{sec:method_iter}

The flowchart of the typical evolutionary algorithm and our \method search method is presented in \cref{fig:evo}. Firstly, an initial \textit{population} containing candidate solutions is generated. Then,  the ``recombination'' process employs \textit{evolutionary operators} (e.g., mutation, crossover) to generate new individuals (offspring) from existing ones, which will be the new population. After that, the \textit{fitness} of each individual in the population is evaluated based on a pre-defined \textit{objective}. Furthermore, based on the fitness values of each individual, a \textit{selection} is applied to filter subpar individuals to acquire the updated population. Such process is repeated until convergence and the last population is acquired.  

Inspired by the process of the evolutionary method, we design an iterative search on both the set- and prompt-level as presented in~\cref{algo:search}. We treat the well-performing prompts as the ``population'', and use the entire text annotations as initialization. For each iteration, we construct a large number of subsets with the prompts within the population $\mathcal{P}'$. It could be viewed as the ``recombination'' process. Then, the generated subsets (offspring) are evaluated using the KD value as the ``fitness''. A novel frequency-based prompt ``selection'' scheme is designed to effectively distinguish promising prompts. The population is updated after the selection process and used in the next iteration. The search algorithm progressively filters the redundant subpar prompts and significantly enhances the search space exploration efficiency, and acquires top-performing subsets with low variance. 


\subsubsection{Frequency-based Prompt Selection}
\label{sec:method_freq}

An effective ``population update'' (or prompt selection) is crucial for the success of the search method. As shown in \cref{fig:bad_cases}, the selected prompts should avoid such corner cases whose standalone KD is high, but interfere with each other, leading to 
poorly performing subsets. We aim to seek prompts that satisfy the property of \textbf{``when combined with any other prompts within the population, it has a higher probability of achieving high KD''}. With the intention of that, we design a frequency-based prompt selection method that selects prompts that appear most frequently in top-performing subsets. In the top-performing sets from a large variety of constructed subsets, the most frequently appearing prompts are likely to satisfy the desired property. The prompts selected this way exhibit high standalone KD, and retain high KD when randomly combined.


\section{Experiments}
\label{sec:exps}

\subsection{Experimental Details}
\textbf{Dataset.} We conducted our experiments on two widely used large-scale datasets: COCO validation dataset (i.e., $val2014$) \cite{COCO} and DiffuisonDB dataset \cite{DiffusionDB}.
In COCO, each image corresponds to several prompts. We randomly select one for each image to result in $40{,}504$ prompts.
DiffusionDB contains 2 million prompts, divided into $2{,}000$ parts. In each part, we uniformly select samples, resulting in a total of $5{,}000$ prompts for our experiments.

\textbf{Metric setting.} We adopt five popular evaluation metrics. Specifically, FID~\cite{FID}, CLIP~\cite{CLIP}, and Aesthetic~\cite{Aesthetic} are employed to assess three distinct aspects including image fidelity, alignment, and aesthetic.
ImageReward~\cite{ImageReward} and HPS~\cite{HPS}) are utilized for evaluating human preference.

\textbf{Model zoo setting.} 
\label{sec:exp-setup}
We construct \textbf{$78$ model configurations} from the StableDiffusion Models~\cite{stable-difffusion} to form the model zoo by considering multiple perspectives. The model zoo comprises \textbf{$12$ variant models} (v1.2, v1.2-6bit quant, v1.2-8bit quant, v1.4, v1.5, small-v1.5, v2.1, dreamlike-photoreal, v1.4-6bit quant, v1.4-8bit quant, v1.5-6bit quant, v1.5-8bit quant) and \textbf{$8$ schedules} (DDIM-10step, DDIM-20step, DDIM-50step, PNDM-10step, PNDM-20step, PNDM-50step, DPM-10step, DPM-20step).
Quantization models are based on the q-diffusion~\cite{q-diffusion} framework, which encompasses both the DDIM and DPM solvers.
For each model, we maintain consistency by fixing seed=$0$ and guidance scale=$7.5$ to ensure fairness in ranking across models.
This model zoo covers various scenarios discussed in \cref{sec:introduction} where users want to evaluate and compare diffusion models, including (1) parameter tuning, such as tuning training iterations/datasets (v1.2, v1.4, v1.5, and dreamlike-photoreal are from different training iterations/datasets of the same model), (2) schedule selection, and (3) validating design choices (e.g., quantization).

\begin{figure*}[h!]
    \centering
    \vspace{-0.1cm}
    \raisebox{3pt}{
    \includegraphics[width=0.95\linewidth]{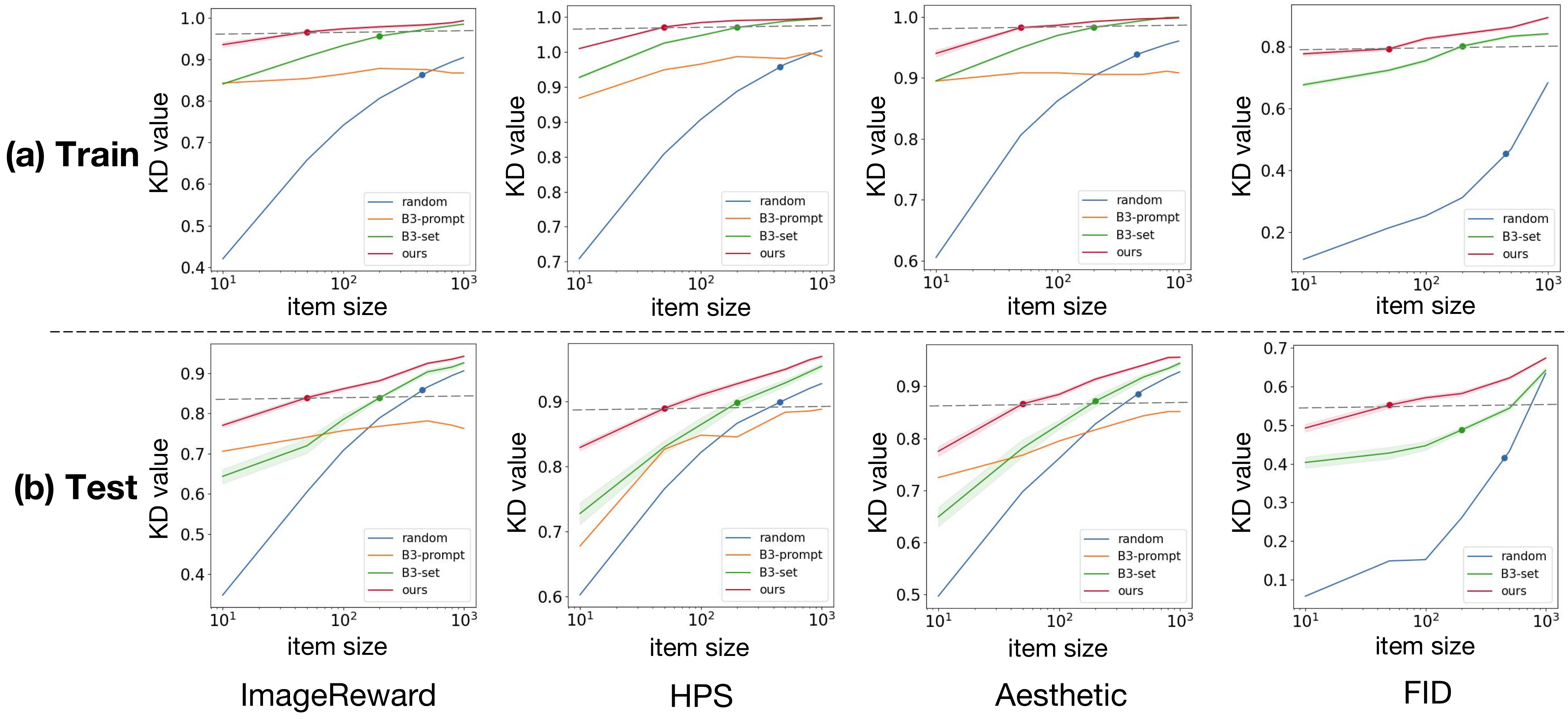}}
    \caption{\textbf{Comparisons with Baseline3 of the Kendall’s Tau of different metrics on COCO dataset for ``Random'' sub-task.} The red, green, and blue dots respectively represent ours with 100 items, B3-set with 200 items, and random sample with 500 items.The shaded area are standard errors. 
    The complete comparison results on the two datasets are in the appendix.}
    \label{fig:compare_result}
    \vspace{-0.2cm}
\end{figure*}

\subsection{Comparison to Baseline Methods}
\label{sec:sub-task}
We design three sub-tasks based on the composition of different models to validate the effectiveness of FlashEval across various settings.
Furthermore, to ensure applicability in diverse scenarios, the representative subset needs to possess the ability to generalize across different models.
%
%
In this regard, we split the model zoo with different partitioning strategies to obtain training models for searching and testing models for assessing generalization capabilities for each sub-task.
The three partitioning ways are as follows: \textbf{random split}, \textbf{across model variances} (different architectures and parameters for training and testing models), \textbf{across schedules} (different solvers and step sizes for training and testing models).
The detailed strategies are in the appendix.

\textbf{FastEval finds subsets with better accuracy-efficiency trade-off.} 
We conduct comparisons with the previously mentioned baseline approaches on each metric for every sub-task.
For the sake of fair comparison, we ensure that the search space for each method, i.e., the total number of candidate subsets, remains consistent ($N=10^{7}$). In addition, each set-based search algorithm is executed 10 times to compute standard errors.
%
Due to FID is only applicable to the evaluation of two distributions and cannot evaluate a single image, it does not support prompt-based searches. Therefore, we just compare \method with set-based methods for FID.

From \cref{tab:COCO} and \cref{tab:diffusionDB}, we can observe that \method can easily achieve KD values around 0.9 on training models when the number of prompts is relatively low.
FlashEval surpasses all methods with equal search space, particularly evident when the item size is small. 
FlashEval at $N'=100$ performs comparably to B3-set at $N'=200$ (Red and green dots in \cref{fig:compare_result}-(a)) across all metrics. 
For prompt-based method, as the number of items in the subset increases, the KD values do not improve, which validates our hypothesis in \cref{sec:method_baseline} that prompt-based method is not suitable for searching representative subset.
In addition, \cref{fig:compare_result}-(a) shows that the commonly used random sampling strategy's performance
drops notably below 500 items (especially on FID).


\textbf{FlashEval generalizes across diverse model setting.} 
We aim for users to utilize the representative set directly on the new models. Thus, we test the generalization performance of FlashEval on \textbf{test models} across three sub-tasks.
We find that FlashEval's set with $N'=5$ surpasses the average KD values of randomly selected sets with $N'=100$ on DiffusionDB (\cref{tab:diffusionDB}). Moreover, the $N'=50$ set identified by FlashEval is on par with the randomly selected $N'=500$ set on COCO ( and \cref{tab:COCO}).
As depicted in \cref{fig:compare_result}-(b), FlashEval exhibits superior generalization across all metrics compared to the baseline methods. Additionally, it notably excels when the representative set has a limited number of prompts.
The dots in \cref{fig:compare_result} also signify that our method achieves comparable performance with $N'=50$ compared to B3-set's $N'=200$ and random sampling's $N'=500$ even for unseen models.
Furthermore, the standard error of FlashEval is relatively smaller compared to B3-set, indicating its higher robustness and stability.

\subsection{Use Cases of \method}

To validate the practicality of FlashEval, we present two examples of downstream use cases here.

\noindent \textbf{Application A}: Model selection from different versions.
\textbf{Experiment settings}: 
%
We adopt a 10-item RS and FlashEval to rank 44 testing models, and the full set (40K) averaged ImageReward is used as ground truth.
We illustrate the superiority of FlashEval using 3 variants of SD1.5 (with DDIM solver \& 20 steps) in testing models: \textit{dreamlike} (fine-tuned), \textit{small-sd1.5} (reduce params), and \textit{weight-8bit} (quantized).
%
\textbf{Result}: \cref{tab:compare_version} provides specific rankings for 3 models. FlashEval with significantly higher KD (0.761) not only reasonably predicts that the dreamlike is better than \textit{small-sd1.5} but also accurately ranks the two efficient models (\textit{small-sd1.5}, \textit{weight-8bit}) that are not intuitively comparable. However, RS failed in both scenarios.
\begin{table}[t]
\centering
\small
\resizebox{0.9\linewidth}{!}{\begin{tabular}{c|ccc}
\toprule
{set}  & {small-sd1.5} & {weight-8bit} & {dreamlike} \\ 
\midrule
RS-10 item (KD=0.286) & 20 & 28 & 25 \\
Ours-10 item (KD=0.761) & 34 &23 & 3  \\
Ground Truth & 39 &25 & 4  \\
\midrule
\end{tabular}}
\caption{\textbf{The ranking of three models among 44 testing models.}}
\vspace{-0.1cm}
\label{tab:compare_version}
\end{table}

\begin{figure}[t]
    \centering
    \includegraphics[width=0.7\linewidth]{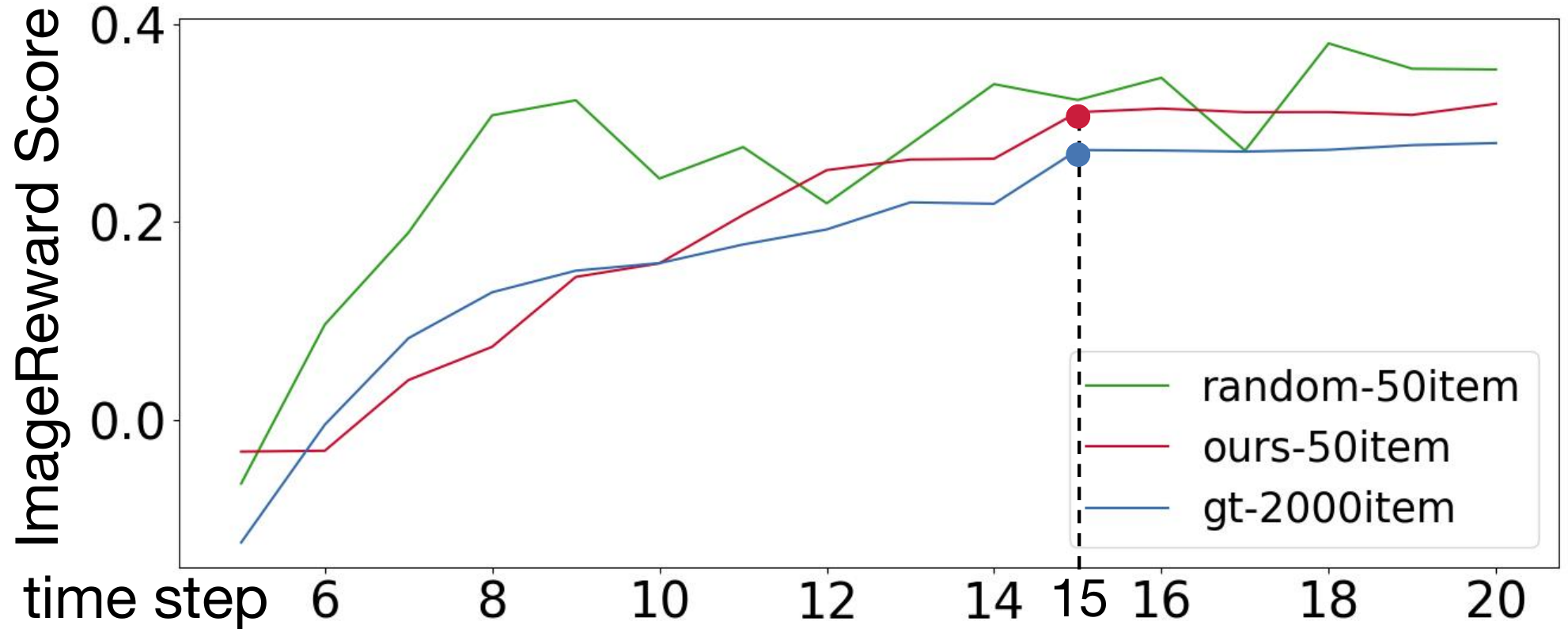}
    \caption{\textbf{The evolution of model scores (sd1.5, DPM solver) with increasing steps.}}
    \vspace{-0.4cm}
    \label{fig:diff_step}
\end{figure}

\label{sec:B}
\noindent \textbf{Application B}: Given a model \& solver, the performance increase saturates as steps increase. Identifying the ``saturate point'' is vital for striking good performance-efficiency trade-off.
\textbf{Experiment settings}: We search the ``saturate point'' by observing that the score remains relatively unchanged in several steps.
\textbf{Result}: \cref{fig:diff_step} shows that FlashEval identifies an accurate saturation point (15) matching ground truth (15).
In contrast, RS fails to find a saturation point in 20 steps due to continuous performance fluctuations.

\subsection{Analysis}

\subsubsection{Ablation Studies}
\hspace{0.3cm} \textbf{Effectiveness of frequency-based selection.} 
To validate the impact of frequency-based selection, the algorithm is modified by selecting prompts with higher KD values, instead of choosing prompts based on their frequency. 


As shown in \cref{tab:ab}, the results are worse than FlashEval, especially on testing models.
In addition, the gap is larger when $N'=10$,
further illustrating the effectiveness of the frequency-based selection module when $N'$ is small.

\begin{figure}[t]
    \centering
    \vspace{-0.2cm}
    \includegraphics[width=0.89\linewidth]{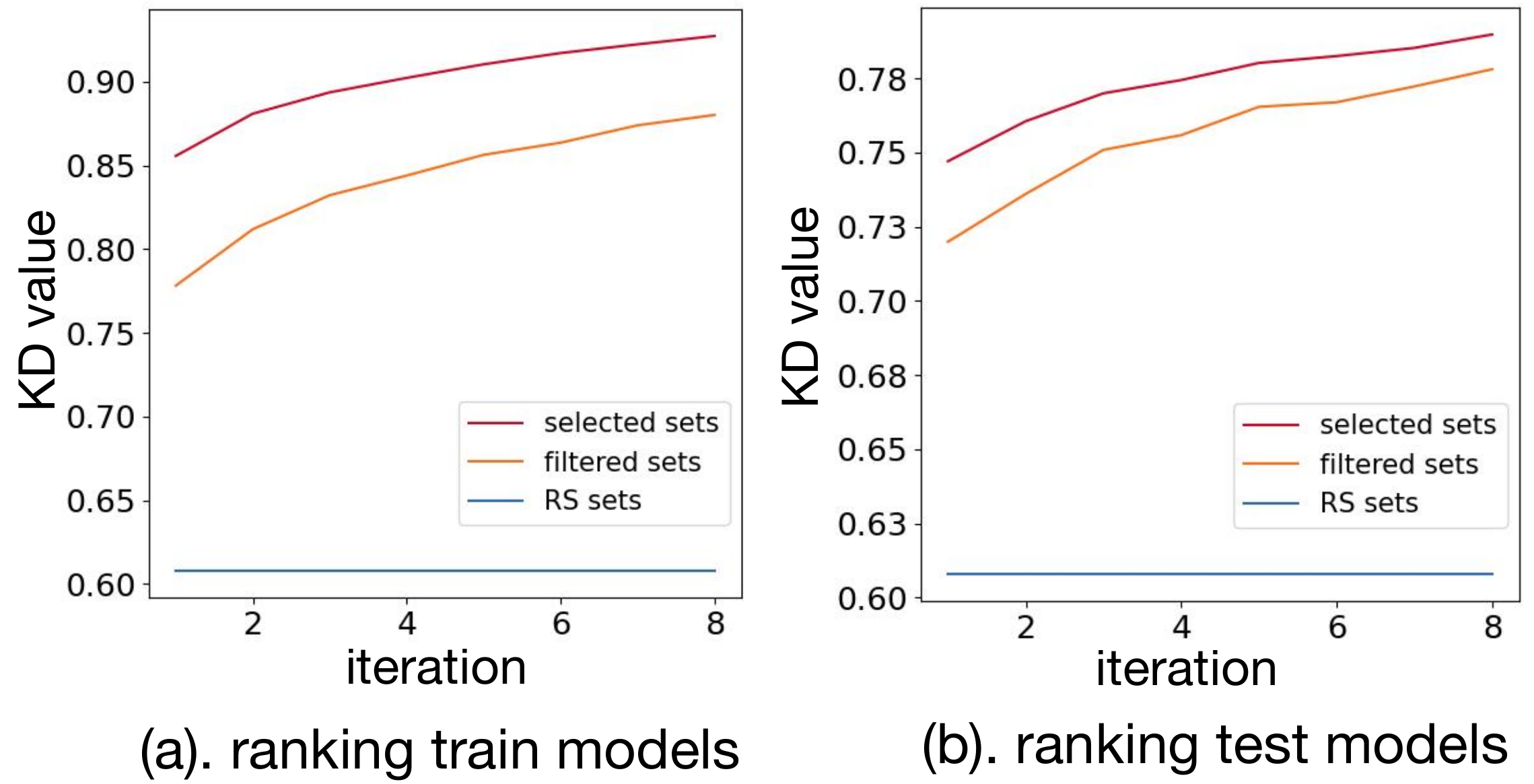}
    \caption{\textbf{The change in KD values as the algorithm progresses through iterations} ($N'=50$, CLIP-Score, COCO, random split).}
    \label{fig:iteraion}
    \vspace{-0.5cm}
\end{figure}

\textbf{Effectiveness of iterative filtering.}
As shown in \cref{tab:ab}, our algorithm demonstrates significantly superior performance on both the training and testing models even without iterative filtering compared to the average performance achieved through random selection.
In addition, from \cref{fig:iteraion}, it is evident that with an increase in the number of iterations, the KD values of the representative set steadily increase. 
In each iteration, the KD values of the filtered set are consistently lower than those of selected set.
The results validate our algorithm's iterative process of gradually filtering out relatively inferior prompts in each iteration.

%
%
%
%
%

\begin{table}[t]
\vspace{-0.4cm}
\begin{center}
\resizebox{0.7\linewidth}{!}{
\begin{tabular}{c|cccc}
\toprule
\multicolumn{1}{c|}{\multirow{2}{*}{Methods}} & \multicolumn{2}{c}{N'=10} & \multicolumn{2}{c}{N'=100} \\ 
\cmidrule(l){2-5} 
\multicolumn{1}{c|}{} & Train & Test & Train & Test \\ 
\midrule
\midrule
RS &  0.355 & 0.142 & 0.706 & 0.692 \\
\midrule
Ours (w/o FS) & 0.876& 0.640 & 0.953 & 0.760 \\
Ours (w/o IF) & 0.821 & 0.667 & 0.839 & 0.812  \\
Ours & \textbf{0.934} & \textbf{0.796} & \textbf{0.966} & \textbf{0.867} \\ 
\midrule
\end{tabular}
}
\end{center}
\vspace{-0.5cm}
\caption{\textbf{Comparison of ablation experiments of CLIP-score on COCO dataset.} ``Ours (w/o FS)'' is our algorithm without frequency-based selection, ``Ours (w/o) IF'' is our algorithm without iteration filtering.}
\label{tab:ab}
\end{table}

\subsubsection{Consistent Superiority of \method}
A high KD value may arise from the correct ranking of only some specific positions (e.g. only the ranking of bottom models are correct).
In such a scenario, a high KD value is irrelevant for users aiming to select the top model.
We add two criteria to verify that FlashEval consistently outperforms RS, rather than focusing on correct ranking at specific positions: (1) \textbf{Top-$K$ Ranking}: KD of the true top-$K$ models. (2) \textbf{Top-$K$ Proportion}: the proportion of true top-$K$ models in predicted top-$K$ models.
\cref{tab:diff_topk} shows that FlashEval consistently outperforms RS across $K$s.

\subsubsection{Search Cost of \method}
The search of \method{} is a one-time cost; the subset found by \method{} can be used in all evaluations of the same dataset.
%
Here, we aim to further reduce this one-time search cost for identifying the representative set for a new dataset. 

Assume that user has a new dataset of size $N_{all}$, and we aim to  
identify a representative set $N'$ of $N_{all}$.
The main computational cost of \method{} is to evaluate the scores of $N_{all}$ samples against all models in the model zoo.
As shown in \cref{sec:sub-task}, random sampling shows poor accuracy with small item size. However, when the number of items is large enough, it is still a cheap approach to achieve reasonable accuracy.
Inspired by this observation, we propose the following two-step approach: 
To find $N'$, we first randomly select a $N_{eval}$ subset ($N'$ < $N_{eval}$ < $N_{all}$) and treat it as the ground-truth when running \method{}. This approach reduces the one-time cost from $\mathcal{O}(N_{all})$ to $\mathcal{O}(N_{eval})$.
$N_{eval}$ is the critical parameter that controls the trade-off between search efficiency and accuracy.
On one hand, $N_{eval}$ should not be too small, as otherwise, it is not representative enough for the true ranking of models (\cref{sec:sub-task}). On the other hand, when $N_{eval}$ is larger, this approach provides less speed-up compared to the vanilla \method{}.

\begin{figure}[t]
    \centering
    \vspace{-0cm}
    \includegraphics[width=0.90\linewidth]{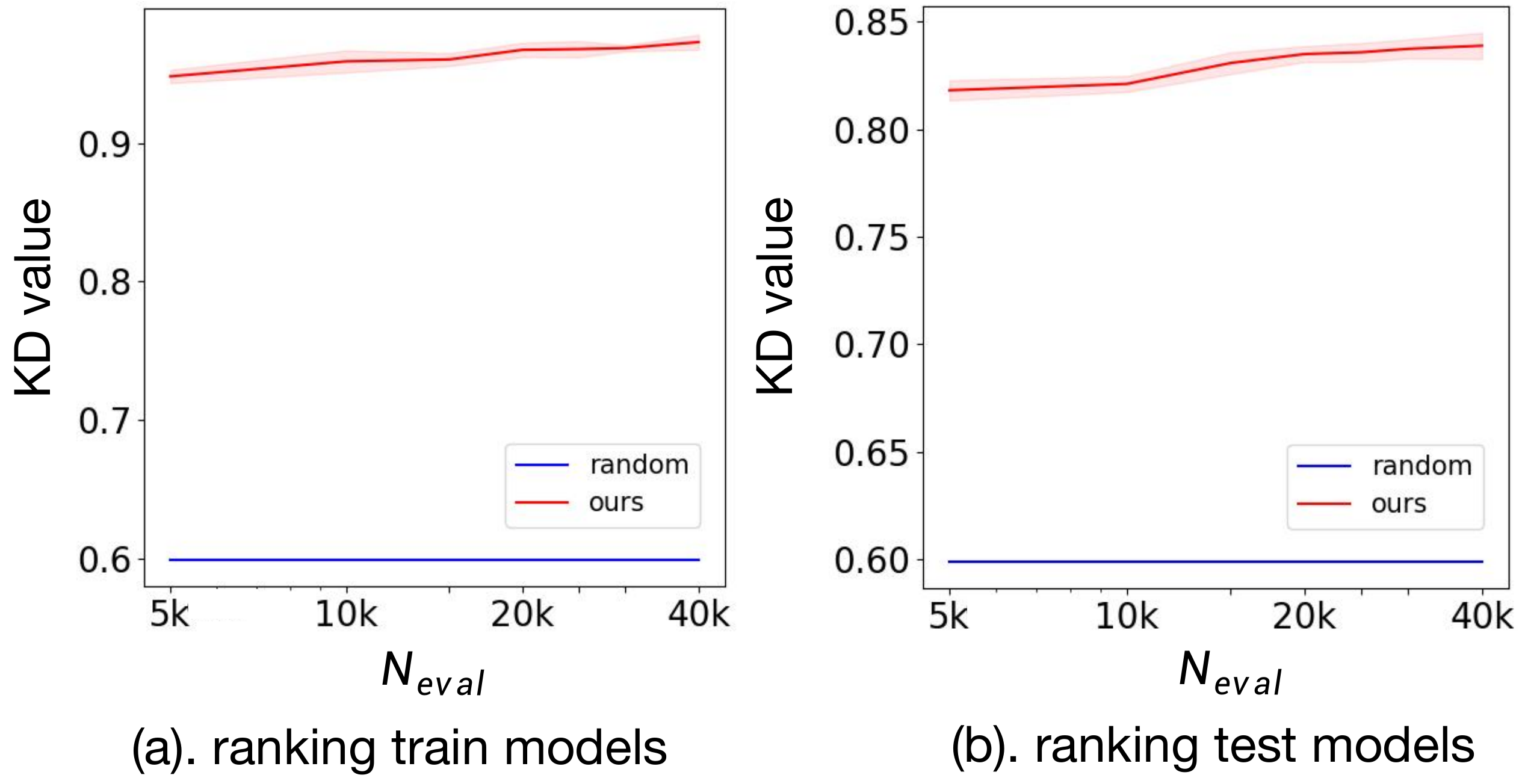}
    \caption{\textbf{The variation of KD values with the increase in $N_{eval}$} ($N'= 50$, CLIP-Score, COCO, random split).}
    \vspace{-0.5cm}
    \label{fig:diff_eval}
\end{figure}

To validate the searched set's effectiveness, we compare the ranking results of $N'$ with the rankings of $N_{all}$.
%
%
As depicted in \cref{fig:diff_eval}, when reducing $N_{eval}$ to $5000$, there is no significant decrease in the results.
In addition, the trade-off in \cref{fig:diff_eval} illustrates the relationship between search cost and performance, aiding users in making their own choices.

%

\begin{table}[t]
\setlength{\abovecaptionskip}{0.2cm}
\vspace{-0.4cm}
\centering
\resizebox{0.65\linewidth}{!}{\begin{tabular}{c|cccc}
\toprule
\multicolumn{1}{c|}{\multirow{2}{*}{Top-$K$}}  & \multicolumn{2}{c}{Top-$K$ Ranking} & \multicolumn{2}{c}{Top-$K$ Proportion} \\ \cmidrule(l){2-5} 
\multicolumn{1}{c|}{} &  RS & Ours & RS & Ours \\ 
\midrule
\midrule
$K=5$ & 0.400 & 0.800 & 40\% & 80\% \\
$K=10$ & 0.520 & 0.810 & 50\% & 80\% \\
$K=20$ & 0.458 & 0.793 & 55\% & 95\% \\
\midrule
\end{tabular}}
\caption{\textbf{Results of 10-item subsets obtained by RS and Ours for various K values.} The total number of testing models is 39.} 
\vspace{0cm}
\label{tab:diff_topk}
\end{table}

\section{Conclusion}
In this paper, we identify the excessive cost and suboptimal quality-speed trade-off of the current diffusion model evaluation.
We propose to address the issues by identifying representative subsets to condense the evaluation datasets and designing a search method to solve the challenging task.
%
%
%
Extensive experiments demonstrate the effectiveness and generalization capability of the representative sets identified by FlashEval.
We hope that FlashEval can assist researchers in selecting an appropriate prompt quantity in diffusion model evaluation, and help accelerate and facilitate the broader diffusion algorithm design.

\noindent\textbf{Acknowledgements.}{
\small
{This work was supported by National Natural Science Foundation of China (No. 62325405, 62104128, U19B2019, U21B2031, 61832007, 62204164), Tsinghua EE Xilinx AI Research Fund, and Beijing National Research Center for Information Science and Technology (BNRist). We thank for all the support from Infinigence-AI.}
}

\clearpage

{\small
\balance
\bibliographystyle{ieee_fullname}
\bibliography{egbib}
}

\end{document}


\title{Supplementary Materials}
\onecolumn

\maketitle
\section{Detailed Description of Textual Feature-Based Baseline Methods}

In this section, we supplement the implementation details of the textual feature space baseline methods (B1/B2) in Sec. 4.1 . 

\subsection{POS (Part of Speech) Tagging Analysis}

\begin{figure*}[h!]
    \centering
    \includegraphics[width=0.98\linewidth]{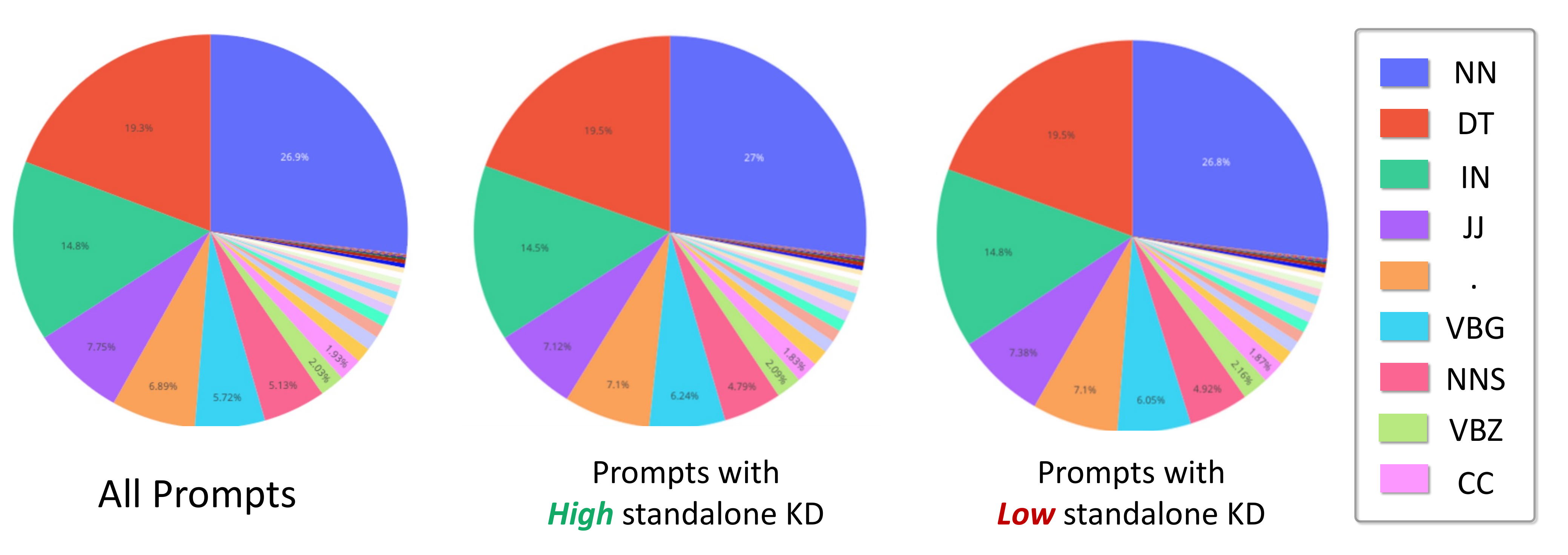}
    \caption{\textbf{Comparison of pos tagging analysis for the entire COCO annotations, 100 item subsets with high/low standalone KD.} }
    \label{fig:pies}
\end{figure*}

The POS (Part of Speech) tagging is a process to mark up the words in text format for a particular part of a speech based on its definition and context. The ``part of speech'' is a typical low-level textual feature, which could potentially reflect the characteristic of the representative textual subset. We conduct the POS tagging analysis on COCO annotations. We use the NLTK's (Nature Language Toolkit) tokenizer and POS tagging tool to analyze the POS distribution. We present the POS that occupies more than 1\% in the legend. The ``NN'' represents ``singular noun'', ``DT'' represents ``determiner'' (e.g., ``my''), ``IN'' represents ``Preposition and conjunction'' (e.g., ``at/in''), ``JJ'' represents ``Adjective'', ``.'' represents the punctuation, ``VBG/VBZ'' represents different forms of verbs, ``NNS'' represents ``Proper and plural noun'', and ``CC'' represents ``conjunction of coordinating'' (e.g., ``that/which'').
As demonstrated in \cref{fig:pies}, the POS distributions of the entire set and subsets of 100 items with high/low standalone KD are \textbf{quite similar}. Therefore, \textbf{it is not feasible to select representative subsets according to this characteristic}.

\subsection{Semantic Class Diversity}
To effectively illustrate the inability to select a representative subset through semantic class diversity, we conduct comprehensive experiments on this matter.
We partition CLIP/BERT features into $10$ clusters using the K-means algorithm, ensuring that prompts in each cluster encompass similar semantic information categories (e.g., humans, cats, etc).
Subsequently, we selected $N'/10$ prompts from each cluster based on the Euclidean distance to the cluster centroid (near or far), or through random selection.
As depicted in \cref{tab:CLIP-feature} and \cref{tab:Bert-feature}, all the methods yields results similar to random sampling.
Therefore, \textbf{ensuring semantic class diversity does not lead to an improvement in performance.}

\begin{table}[h]
\centering
\begin{tabular}{c|cccc}
\toprule
\multicolumn{1}{c|}{\multirow{2}{*}{Methods}}  & \multicolumn{2}{c}{$N'=10$} & \multicolumn{2}{c}{$N'=100$} \\ \cmidrule(l){2-5} 
\multicolumn{1}{c|}{} & Train & Test & Train & Test \\ 
\midrule
\midrule
RS & 0.355 & 0.346 & 0.706 & 0.692 \\
\midrule
Near Centroid & 0.455 & 0.451 & 0.474 & 0.570  \\
Far Centroid & 0.551 & 0.539 & 0.617 & 0.713 \\
RS-each cluster & 0.341 & 0.313 & 0.617 & 0.682 \\
\midrule
\end{tabular}
\caption{\textbf{Comparisons of the Kendall's Tau of CLIP-Score for subsets acquired by clustering CLIP features.} } 
\label{tab:CLIP-feature}
\end{table}

\begin{table}[h]
\centering
\begin{tabular}{c|cccc}
\toprule
\multicolumn{1}{c|}{\multirow{2}{*}{Methods}}  & \multicolumn{2}{c}{$N'=10$} & \multicolumn{2}{c}{$N'=100$} \\ \cmidrule(l){2-5} 
\multicolumn{1}{c|}{} & Train & Test & Train & Test \\ 
\midrule
\midrule
RS & 0.355 & 0.346 & 0.706 & 0.692 \\
\midrule
Near Centroid & 0.057 & 0.335 & 0.490 & 0.540 \\
Far Centroid & 0.285 & -0.116 & 0.659 & 0.545 \\
RS-each cluster & 0.228 & 0.251 & 0.686 & 0.651 \\
\midrule
\end{tabular}
\caption{\textbf{Comparisons of the Kendall's Tau of CLIP-Score for subsets acquired by clustering Bert features.} } 
\label{tab:Bert-feature}
\end{table}

\begin{algorithm}[t]
    \caption{Greedy Search Algorithm}
    \resizebox{0.9\linewidth}{!}{ 
    \begin{minipage}{0.99\linewidth}
    \textbf{Input:}  Whole textual dataset $\mathcal{P} = \{p_i | i=1,...,N\}$, The corresponding features of dataset $\mathcal{F} = \{f_i | i=1,...,N\}$, The mean $\mu_{all}$ and covariance $\Sigma_{all}$ of multivariate Gaussian distribution of $\mathcal{F}$;
  
    \textbf{Ouput:} Subsets $ \hat{\mathcal{P}}$ of size $N'$;
    
    Random Initialize $\hat{\mathcal{P}} \gets {p_k}, \hat{\mathcal{F}} \gets {f_k}  (1 \leq k \leq N)$;
    
    \For{$i \gets 2$ to $N'$}{
        $index = INF, D_{min} = INF$;
        
        \For{$f_j $ in $\mathcal{F}-\hat{\mathcal{F}}$}{
            $\mathcal{F}_{j} = \hat{\mathcal{F}} \cup f_j$;
            
            Calculate $\mu_{j}$ and $\Sigma_{j}$ of $\mathcal{F}_{j}$;

            \tcp{KL distance between $\mathcal{F}_j$ and $\mathcal{F}$}
            
            $D_j = KL (\mu_{j}, \Sigma_{j}, \mu_{all}, \Sigma_{all})$

            \If{$D_{min} > D_j$}{     
                $index = j, D_{min} = D_j$;
                }
        }

        $ \hat{\mathcal{P}} = \hat{\mathcal{P}} \cup p_{index}, \hat{\mathcal{F}} = \hat{\mathcal{F}} \cup f_{index}$;
  }  
  
  \Return $\hat{\mathcal{P}}$ 
  
  \end{minipage}
  }
  \label{algo:greedy}
\end{algorithm}

\subsection{Statistical Property Similarity}
\begin{figure*}[h!]
    \centering
    \includegraphics[width=0.6\linewidth]{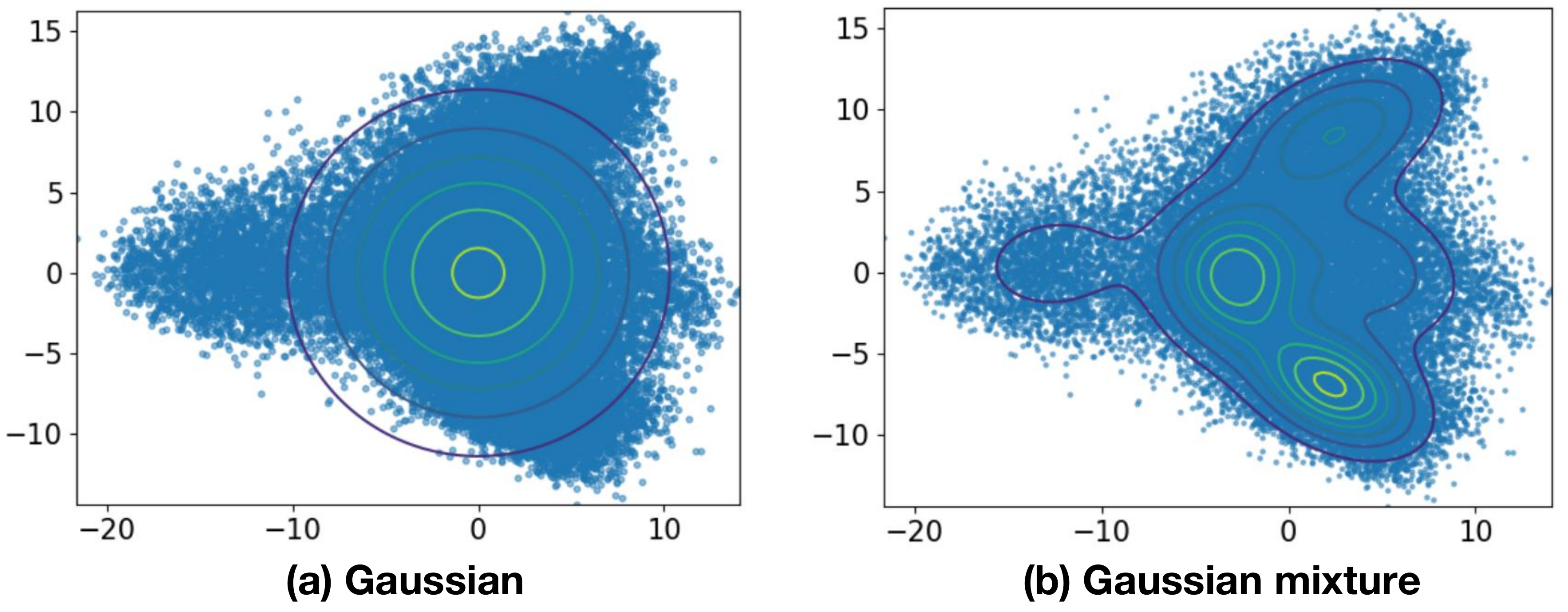}
    \caption{\textbf{T-SNE Feature Visualization of Gaussian and Gaussian Mixture (\#component = 5) distributions fitted to all features.} }
    \label{fig:GMM}
\end{figure*}

As shown in Algo.~\ref{algo:greedy}, we design a greedy search algorithm that selects the prompt that minimizes the KL divergence between the CLIP feature distribution of the subset and the entire set in each iteration.
%
%
Since the algorithm requires computing the KL divergence, the feature distribution is assumed to be a multivariate Gaussian distribution.
To better fit the feature distribution, we expand the Gaussian distribution into the Gaussian mixture distribution.
It can be depicted in \cref{fig:GMM}, when the Gaussian Mixture distribution contains five Gaussian functions, it effectively fits the features.
Then, we run the algorithm in each Gaussian function, identifying $N'/5$ prompts in each function (a total of $N'$ prompts).
We show the results of different items in \cref{tab:gmm}. \textbf{It was evident that matching the statistical property similarity does not improve the ranking outcome}.

\clearpage

\begin{table}[h!]
\centering
\small
\begin{tabular}{c|cccccccc}
\toprule
\multicolumn{1}{c|}{\multirow{2}{*}{Methods}}  & \multicolumn{2}{c}{$N'=100$} & \multicolumn{2}{c}{$N'=200$} & \multicolumn{2}{c}{$N'=500$} & \multicolumn{2}{c}{$N'=1000$} \\ \cmidrule(l){2-9} 
\multicolumn{1}{c|}{} & Train & Test & Train & Test & Train & Test & Train & Test\\ 
\midrule
\midrule
RS & 0.706 & 0.692 & 0.784 & 0.763 & 0.857 & 0.829 & 0.896 & 0.867 \\
\midrule
Gaussian & 0.714 & 0.696 & 0.764 & 0.791 & 0.873 & 0.847 & 0.913 & 0.872 \\
Gaussian Mixture & 0.565 & 0.420 & 0.809 & 0.770 & 0.799 & 0.802 & 0.881 & 0.880 \\
\midrule
\end{tabular}
\caption{\textbf{Comparisons of the Kendall's Tau of CLIP-Score for subsets acquired by statistical property similarity.} } 
\label{tab:gmm}
\end{table}

\section{FlahEval's Capability of Estimating Model Scores}
\begin{figure*}[h!]
    \centering
    \includegraphics[width=0.8\linewidth]{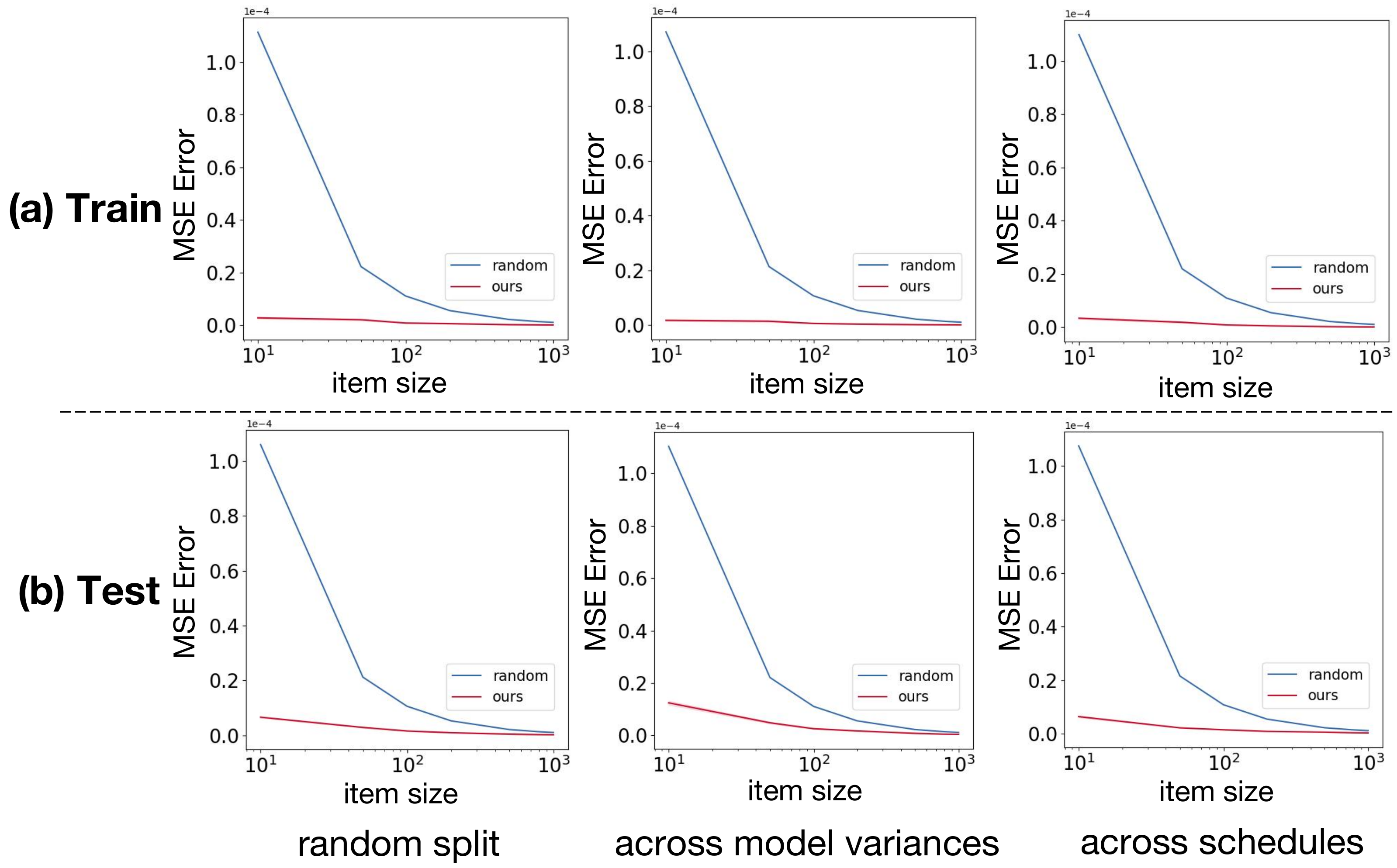}
    \caption{\textbf{Comparisons with random sampling of the estimation quality for CLIP-Score on COCO dataset.} Small MSE Error is better.} 
    \label{fig:score-coco}
\end{figure*}

\begin{figure*}
    \centering
    \includegraphics[width=0.8\linewidth]{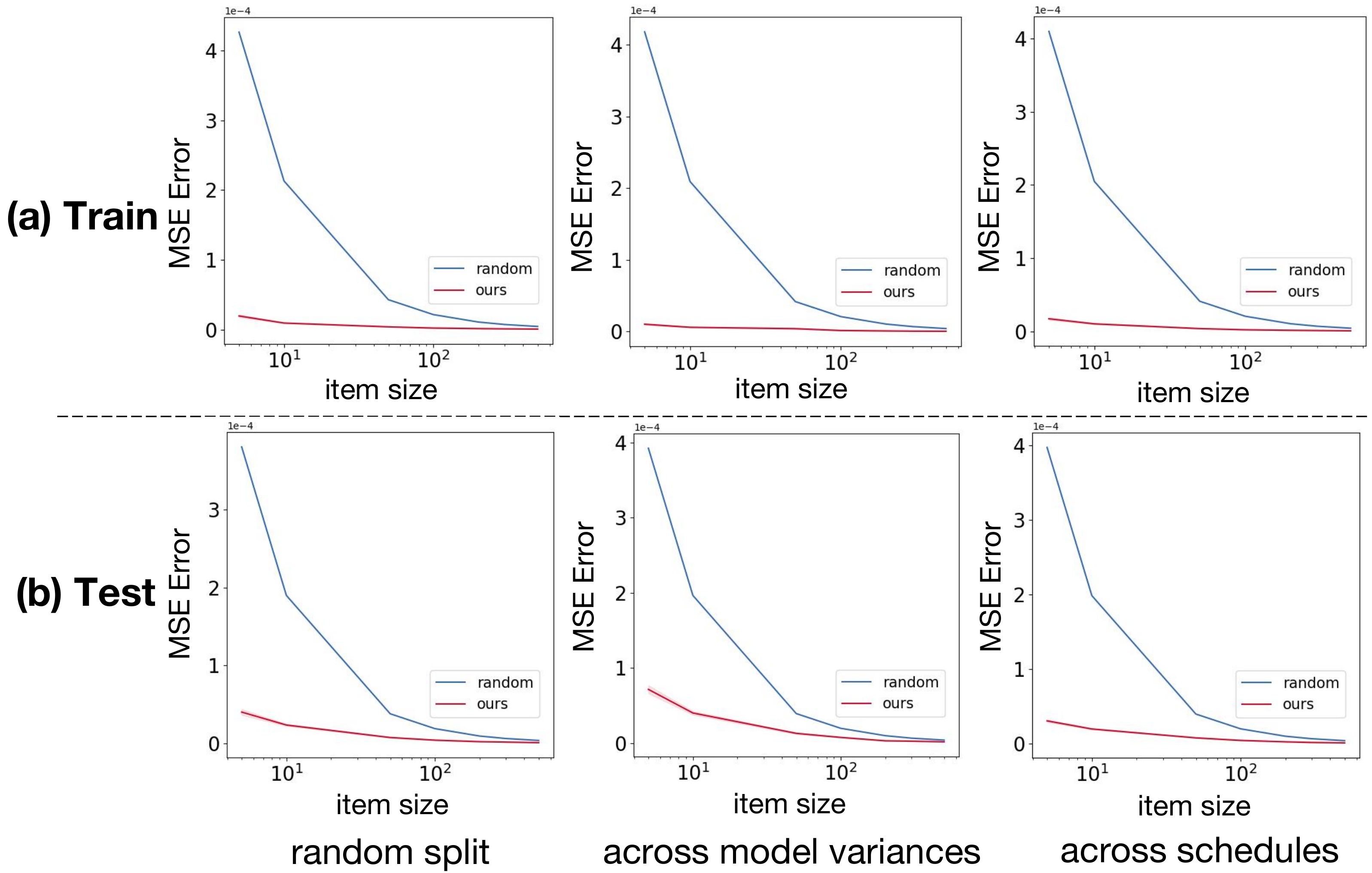}
    \caption{\textbf{Comparisons with random sampling of the estimation quality for CLIP-Score on DiffusionDB dataset.} Small MSE Error is better.} 
    \label{fig:score-diffusionDB}
\end{figure*}

In some scenarios, users may require the exact metric scores of models rather than the relative model rankings.
\method also supports fitting model scores using a small subset of $N'$ \textbf{by searching for the mean square error}.
As shown in \cref{fig:score-coco} and \cref{fig:score-diffusionDB}, we demonstrate a comparison of our method against random sampling for model scores prediction.
It is evident that our method can accurately estimate model scores even with smaller $N'$ and it generalizes well to testing models across three sub-tasks.

\clearpage

\section{Analysis of Alternative Approaches to Acquire Representative Subsets}
\label{sec:error}

In the main paper, we recognize that ``some subsets with high KD has large estimation error'', which reflects such sets lack precise estimation of the performance. Therefore, an intuitive question arises ``could we use the estimation error as the metric for recognizing representative subsets?''. We could measure the ``estimation error'' by the MSE (Mean Squared Error) loss between the ground-truth performance (averaged on the entire set) and the subset's performance $S^{’}_j = \sum_{i \in \mathcal{I}} \frac{1}{N'} \mathcal{S}_{ij}$  for each model setting. It could be described as follows ($N_m$ is the number of model settings):

\begin{equation}
  \begin{split}
  \varOmega_{EA} = \frac{1}{N_{m}} \sum_{i=1}^{N_m}(\hat{\mathcal{S}_j} - S^{’}_j)^2.
  \end{split}
  \label{equ:mse_loss}
\end{equation}

In this section, we investigate this assumption. We discuss some intuitive alternative approaches to acquire representative subsets based on estimation error (``prompts with least estimation error'',``sets with least estimation error'') and demonstrate their failure. Furthermore, We analyze the underlying reason for their failure and why the FlashEval search method is necessary. 

\begin{figure*}[h!]
    \centering
    \includegraphics[width=0.5\linewidth]{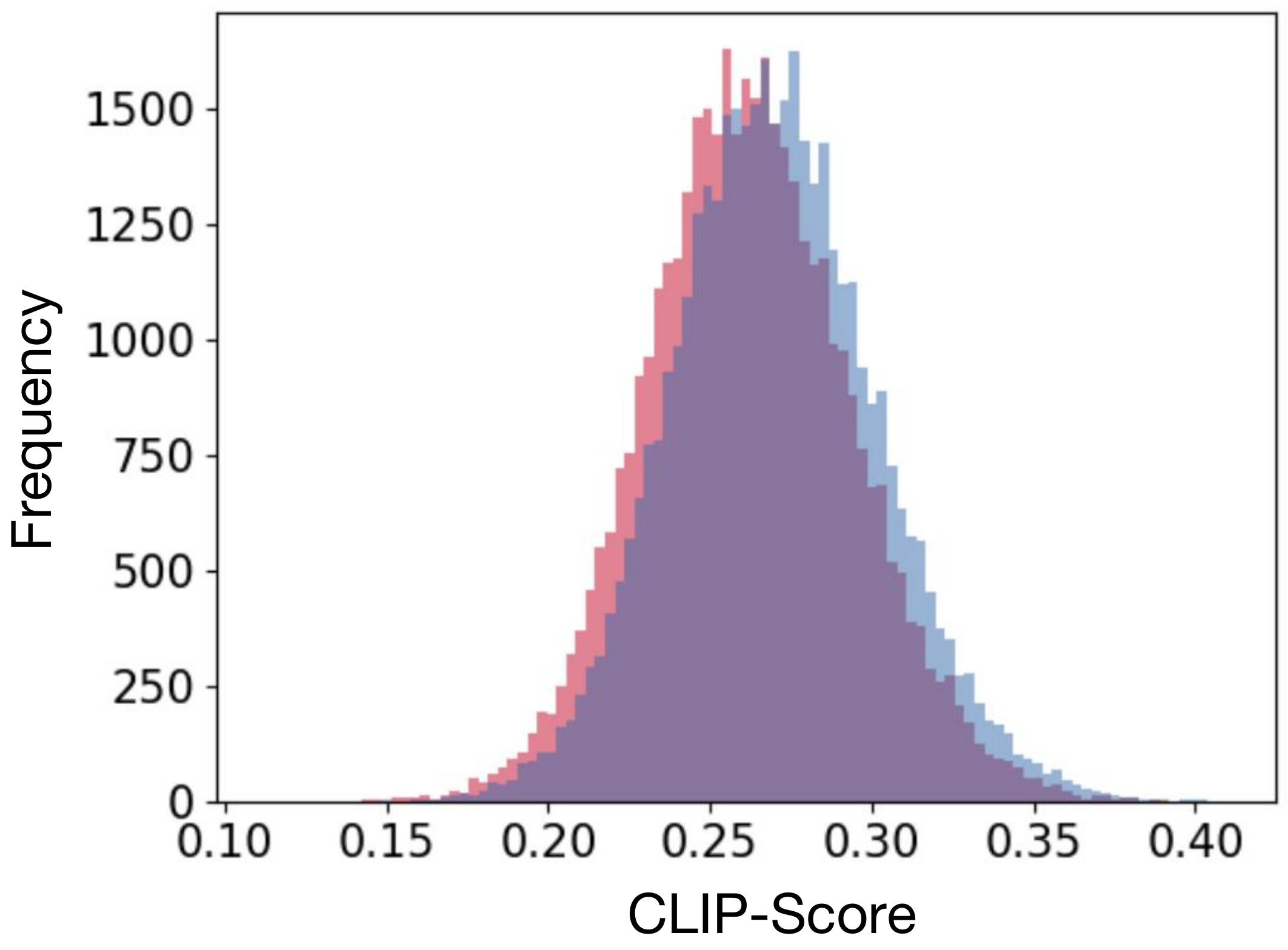}
    \caption{\textbf{Visualization of CLIP-Score of all prompts for two models.} The x-axis representing CLIP-Score values and the y-axis indicating the count of prompts achieving that value.}
    \label{fig:score_fre}
\end{figure*}

As shown in \cref{fig:score_fre}, we visualize the scores of all prompts in COCO dataset across two randomly selected models.
It is noticeable that scores obtained by prompts for a certain model generally follows a Gaussian distribution. The mean of the gaussian distribution is the ``ground-truth'' model performance evaluation across the entire set. Therefore, if there exists prompts/subsets that could correctly approximate the mean of the gaussian distribution for all models, using the prompt/subsets for evaluation will have both high KD values and low estimation error. 

\textbf{Choose prompts with lowest estimation errors.} To validate this existence of such prompts, we separately select $50$ prompts with high KD values and $50$ prompts with low estimation errors, examining their performance in the other criterion.
From \cref{tab:kd_error}, we observe that prompts with high KD values have large estimation errors, while prompts with low estimation errors also exhibit small KD values. We conclude that \textbf{single prompt could not simultaneously satisfy high KD and estimation error, which means that ``prompt that approximates the mean for all models'' does not exist}. Therefore, simply choosing prompts with least estimation error to construct subsets is not feasible. 

\begin{figure*}[h!]
    \centering
    \includegraphics[width=0.5\linewidth]{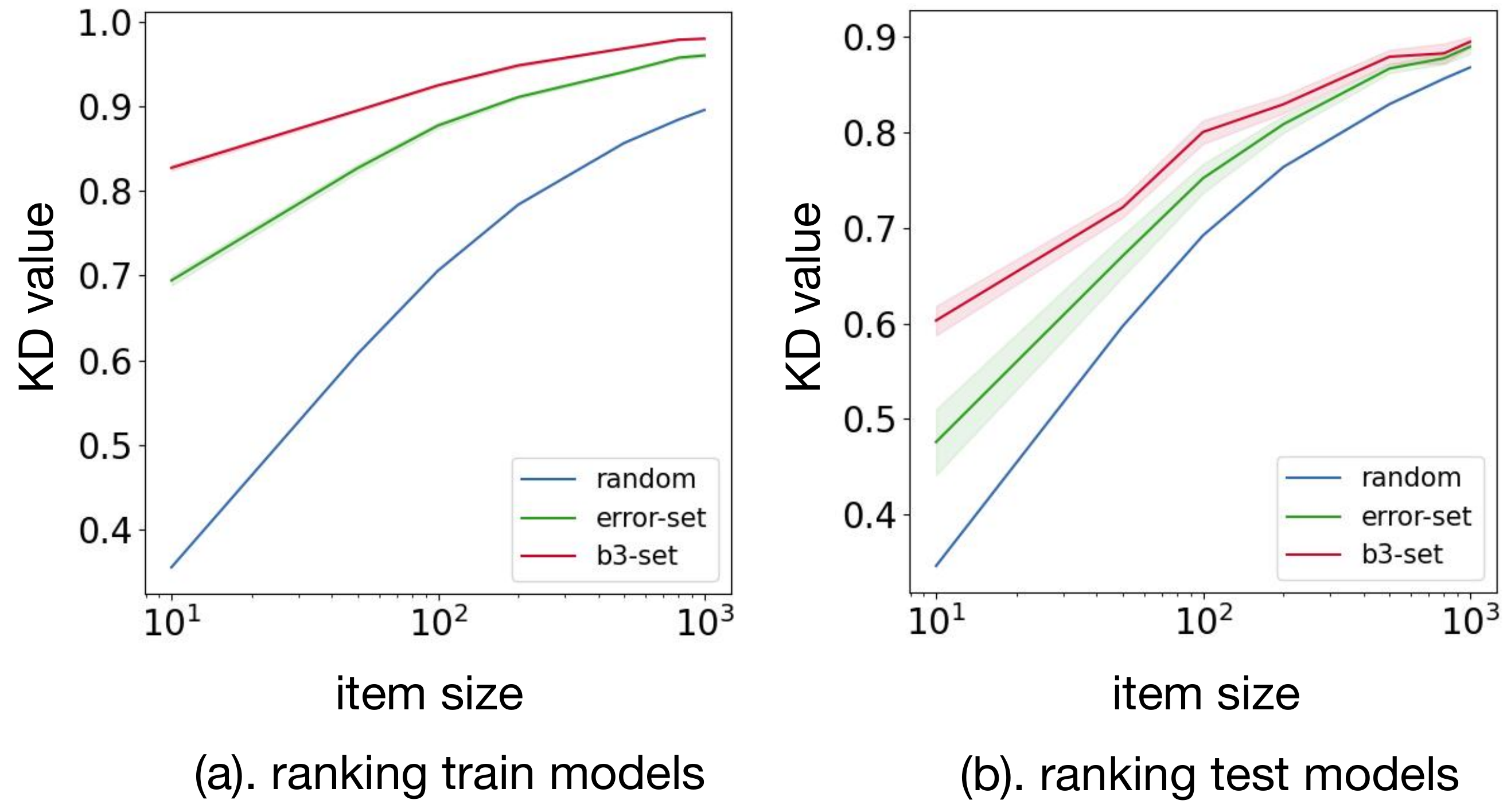}
    \caption{\textbf{Comparisons of set-based baselines using Kendall’s Tau for CLIP-Score on COCO dataset.} The shaded area are standard errors.}
    \label{fig:mean_set}
\end{figure*}

\textbf{Choose subsets with lowest estimation errors.} Similarly, we design experiments to validate the existence of the subsets that could precisely estimate the performance of each model. We generate $10{,}000{,}000$ random samplings at various $N'$ values, and select a set with the highest KD value and a set with the lowest estimation error. From \cref{fig:mean_set}, it is evident that the set we discovered with the lowest error reaches only a KD value of roughly around $0.7$ in training models when $N'=10$. However, the subset with the highest KD value which b3-set discovered can achieve KD values above $0.8$. 
Therefore, simply choosing subsets with least estimation error to construct subsets is also not feasible.

%
%
%

\begin{figure*}[h!]
    \centering
    \includegraphics[width=0.9\linewidth]{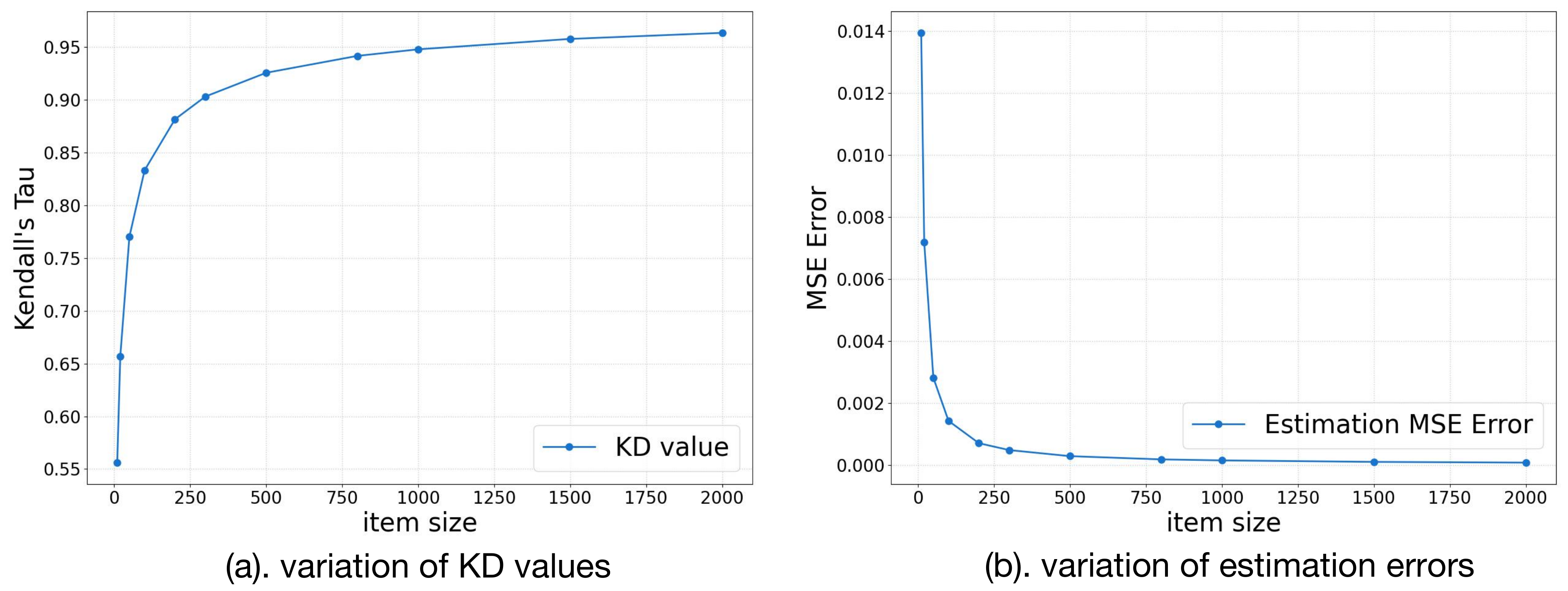}
    \caption{\textbf{The variation of average KD values and estimation errors with increasing $N'$.} The blue dots represent the average values obtained from $100{,}000$ random samplings at the current $N'$.} 
    \label{fig:trade_off}
\end{figure*}

\textbf{The underlying reason for their failure:}
As discussed above, in \cref{fig:mean_set}, we can see that when $N'$ is small, the subset chosen based on estimation error performs poorly, indicating a lack of consistency between KD values and estimation errors. However, as $N$ increases, the subsets acquired by ``error-set'' achieves relatively high KD.
To further validate the above findings, we present the KD values and estimation errors with respect to $N'$ in \cref{fig:trade_off}. We discover that the MSE error decreases sharply with $N'<500$. 
Such phenomenon reveals the potential reason for ``high KD and low estimation error could not be simultaneously satisfied for a smaller $N'$''. When the $N'$ is small, the estimation error is still relatively large (larger than 0.002), in which case, relatively lower estimation error does not guarantee higher KD. However, as $N'$ increases, the estimation error is sufficiently small to ensure high correlation between lower estimation error and higher KD.

%
%

\begin{table}[h!]
\begin{center}
\begin{tabular}{c c c}
\hline
\toprule[1pt]
Methods & Kendall's Tau\(\uparrow\) & Estimation Error\(\downarrow\)  \\
\midrule
All prompts mean & \cellcolor{red!15} 0.269/0.201 & \cellcolor{red!15} 0.142/0.138  \\
\midrule
prompts with low errors & \cellcolor{red!15} 0.391/0.245 & \cellcolor{green!15}  0.052/0.062 \\
prompts with high KD values & \cellcolor{green!15} 0.678/0.463 & \cellcolor{red!15} 0.117/0.117 \\
\bottomrule[1pt]
\end{tabular}
\end{center}
\caption{\textbf{The average performance of the top $50$ prompts in one criterion on the other criterion for training/testing models.} The ``All prompts mean'' represents the averaged value for all independent prompts.}
\label{tab:kd_error}
\end{table}

\clearpage

\section{Qualitative Results of FlashEval Acquired Subsets}

\begin{figure*}
    \centering
    \includegraphics[width=1.0\linewidth]{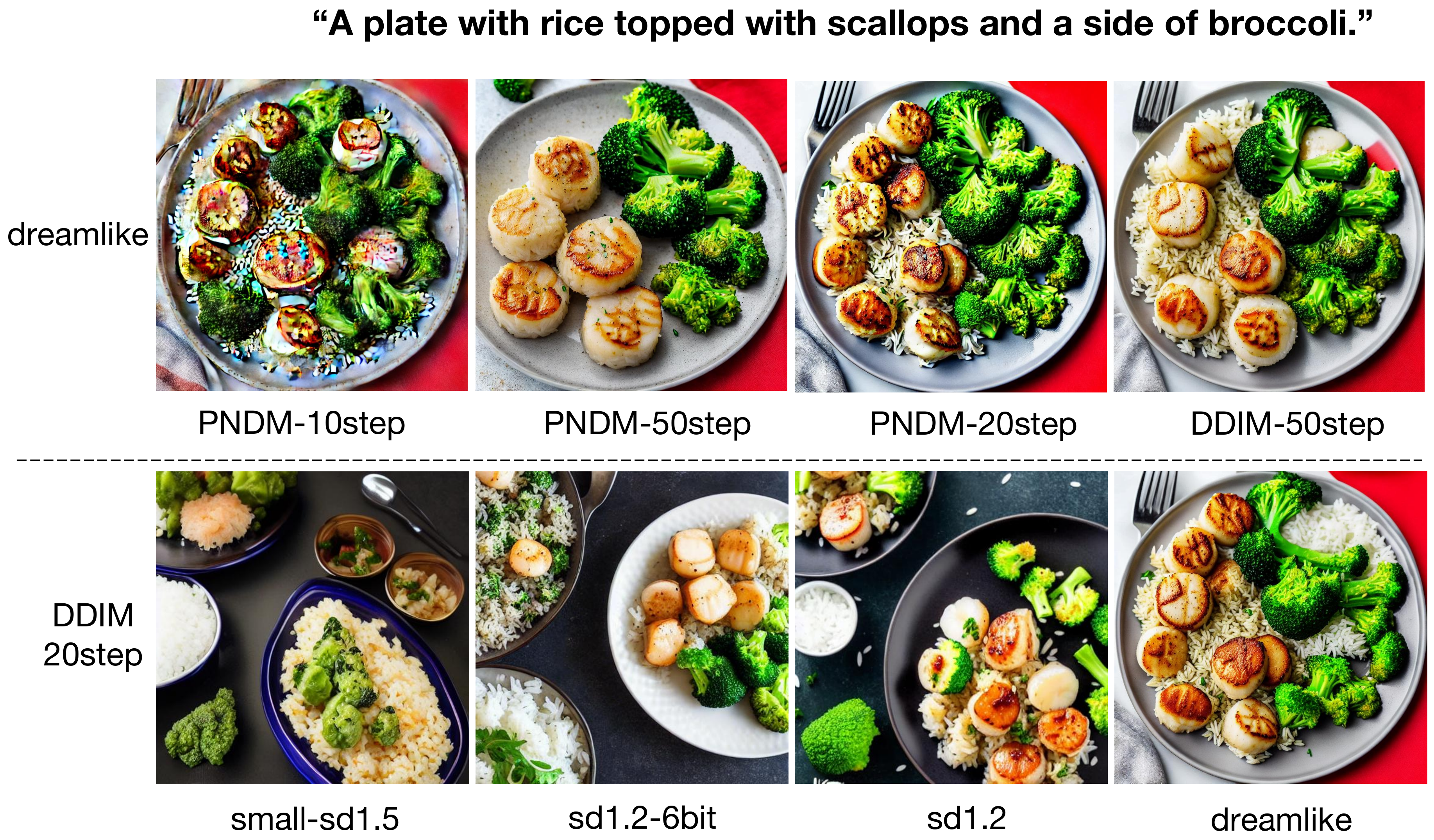}
    \caption{\textbf{Visualization of images generated by the prompt from the representative set of CLIP-Score across various models.} The first row represents variations among different schedules in the same model, while the second row depicts variances among different model architectures or parameters under the same schedule.}
    \label{fig:photo-CLIP}
\end{figure*}

\begin{figure*}[h!]
    \centering
    \includegraphics[width=1.0\linewidth]{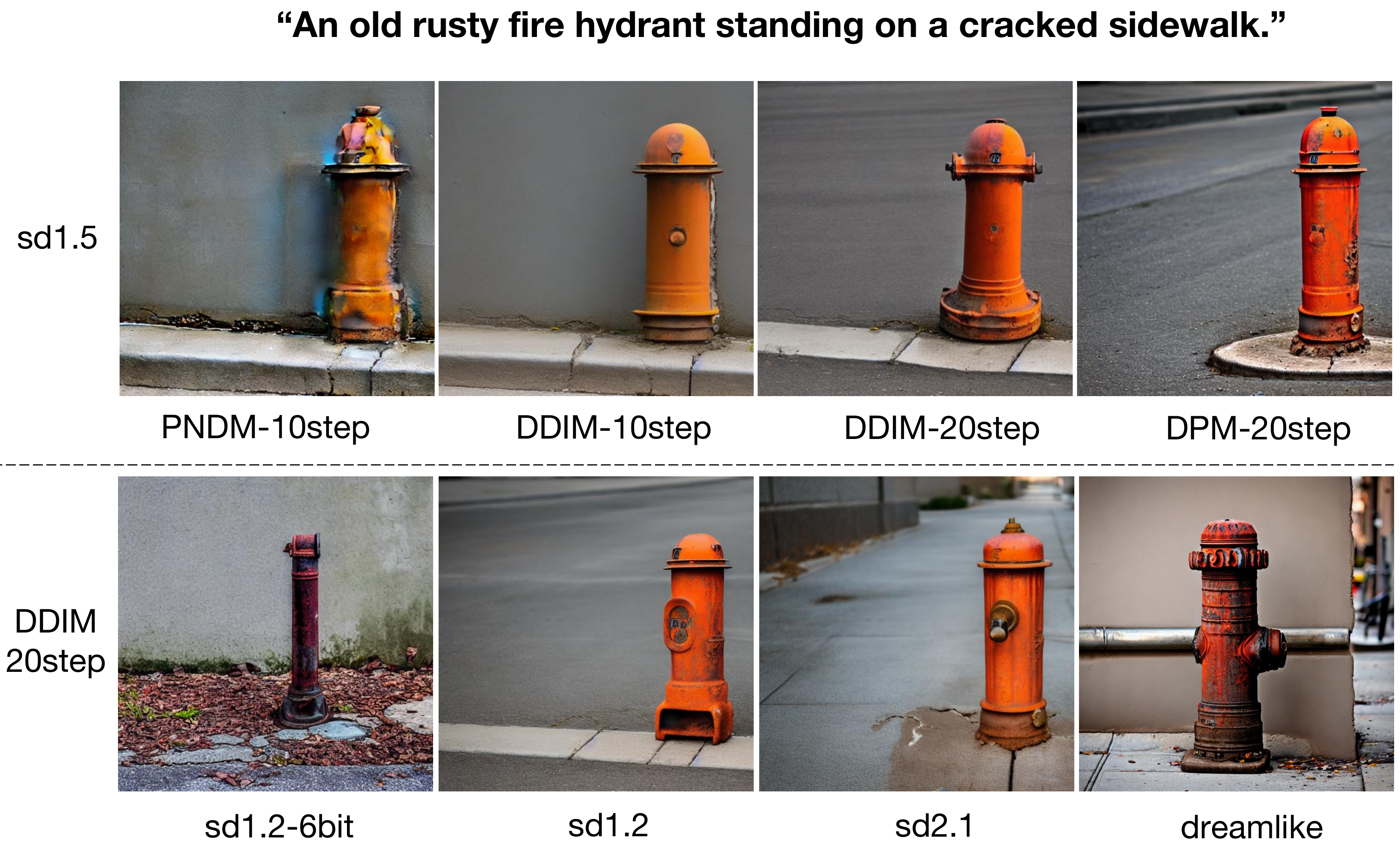}
    \caption{\textbf{Visualization of images generated by the prompt from the representative set of ImageReward across various models.} The first row represents variations among different schedules in the same model, while the second row depicts variances among different model architectures or parameters under the same schedule.}
    \label{fig:photo-ImageReward}
\end{figure*}

In this section, we present and analyze the FlashEval searched textual subset, to demonstrate the searched subset effectively distinguishes different model settings. We choose the $N'=10$ item subset for the COCO annotations.
\cref{tab:prompt-CLIP} and \cref{tab:prompt-ImageReward} respectively outline the prompts included in the representative set for CLIP and ImageReward.
Additionally, we show the images generated by prompts across different models in \cref{fig:photo-CLIP} and \cref{fig:photo-ImageReward}. The upper row displays the generated images under different schedules of the model ``dreamlike'', sorted by the ground-truth metric values (the right has higher metric values). As could be seen, the image visual quality for this prompt (“A plate with rice topped with scallops and a side of broccoli.”) aligns with the ground-truth performance. It verifies that the searched representative subsets contain prompts correctly reflecting the model performance. Similarly, the lower row demonstrates that when using the same schedule, the prompt could also correctly rank different models. Aside from the CLIP-Score, we also present the generated images picked by human-preference-based metric ImageReward in \cref{fig:photo-ImageReward}, similar alignment could be witnessed.

%
%

\begin{table}[h!]
\centering
\begin{tabular}{c|c}
\toprule
\multicolumn{1}{c|}{item size}  & prompts in the representative set \\ 
\midrule
\multirow{10}{*}{$N'=10$} & "A plate with rice topped with scallops and a side of broccoli." \\
 & "A decorative Asian lantern sculpture in the garden with flower ornaments." \\
 &  "A man is riding his skateboard down the road."\\
 &  "A close-up of a plate of food that has been eaten."\\
 &  "A small pizza in the middle of a table."\\
 &  "A white plate topped with a hot dog, french fries and condiments."\\
 &  "A black and white cat sitting on top of a pile of clothes."\\
 &  A giraffe and a zebra are on the grass near trees $\&$ cars. \\
 &  "A busy bus station with ramp going downstairs."\\
 &  "Heavily loaded green truck with passengers on back in roadway in urban area."\\
\midrule
\end{tabular}
\caption{\textbf{Prompts in the representative set of CLIP-Score when $N'=10$ on COCO dataset.}} 
\label{tab:prompt-CLIP}
\end{table}

\begin{table}[h!]
\centering
\begin{tabular}{c|c}
\toprule
\multicolumn{1}{c|}{item size}  & prompts in the representative set \\ 
\midrule
\multirow{10}{*}{$N'=10$} & "Two giraffes standing around on the grassy plains." \\
 & "there is someone holding a remote in there hand " \\
 & "A man adjusts numbers at a tennis match." \\
 & "A skateboarder catches major air during this stunt." \\
 & "A little boy in a yellow shirt feeding a giraffe.  " \\
 & "An old building with two rusty, very old pick up trucks parked in front." \\
 & "A little girl sitting on top of a bed next to a lamp." \\
 & "An old rusty fire hydrant standing on a cracked sidewalk." \\
 & "A man driving a car across an airport runway." \\
 & "The german shepherd is guarding the front of the barred door," \\
\midrule
\end{tabular}
\caption{\textbf{Prompts in the representative set of ImageReward when $N'=10$ on COCO dataset.}} 
\label{tab:prompt-ImageReward}
\end{table}

\clearpage

\section{Verification of the Generelization Ability of FlashEval}

As mentioned in Table.2 and Table.3 in the main paper, we design 3 distinct model settings train-test split to verify the generalization ability of FlashEval. They act as 3 diverse subtask to test the generalization ability across random settings, across models, and across schedules. In this section, we firstly provide a detailed description of the 3 train-test split. Then, we present the results on these subtasks on COCO annotations (\cref{fig:CLIP_coco_com}, \cref{fig:ImageReward_coco_com}, \cref{fig:HPS_coco_com},  \cref{fig:Aesthetic_coco_com}, and \cref{fig:FID_coco_com}) and DiffusionDB datasets (\cref{fig:CLIP_diffusionDB_com}, 
 \cref{fig:ImageReward_diffusionDB_com}, \cref{fig:HPS_diffusionDB_com}, \cref{fig:Aesthetic_diffusionDB_com}, and \cref{fig:FID_diffusionDB_com}).

\subsection{Detailed List of Models}
In \cref{tab:random-split,tab:across-model,tab:across-schedules}, we present detailed lists of models encompassed by both the training and testing models for each sub-task. We conduct the search on the training models and validate the ranking correlation on the testing models to verify the generalizaiton ability of FlashEval. The ``random split'' randomly splits the model setting zoo into two halves, the models of the training and testing splits have diverse model variants, solver, and timesteps. The ``across model variants'' uses only 4 models (SD-1.4, SD-1.5, SD-1.5-6bit,SD-2.1) as the training models, and the rest as the test models. The ``across schedules'' uses the ``PNDM-10'', ``DDIM-10'', ``DDIM-20'', ``DDIM-50'' as the train models. 

\begin{table}[h!]
\centering
\begin{tabular}{ccc|ccc}
\toprule
\multicolumn{3}{c|}{training models} & \multicolumn{3}{c}{testing models} \\ \cmidrule(l){1-6}
model & solver & step & model & solver & step\\ 
\midrule
\midrule
stablediffusion1.2 & PNDM & 10 &
stablediffusion1.2 & PNDM & 20  \\
stablediffusion1.2 & DDIM & 10 &
stablediffusion1.2 & PNDM & 50 \\ stablediffusion1.2 & DDIM & 20 &
stablediffusion1.2-6bit & DDIM & 10 \\
stablediffusion1.2 & DDIM & 50 &
stablediffusion1.2-8bit & DDIM & 10 \\
stablediffusion1.2 & DPM & 10 &
stablediffusion1.2-8bit & DPM & 10 \\
stablediffusion1.2 & DPM & 20 &
stablediffusion1.2-8bit & DPM & 20 \\
stablediffusion1.2-6bit & DDIM & 20 &
stablediffusion1.4 & PNDM & 20 \\
stablediffusion1.2-6bit & DDIM & 50 &
stablediffusion1.4 & PNDM & 50 \\
stablediffusion1.2-6bit & DPM & 10 &
stablediffusion1.4 & DDIM & 20 \\
stablediffusion1.2-6bit & DPM & 20 &
stablediffusion1.4 & DPM & 20 \\
stablediffusion1.2-8bit & DDIM & 20 &
stablediffusion1.4-6bit & DDIM & 20 \\
stablediffusion1.2-8bit & DDIM & 50 &
stablediffusion1.4-6bit & DPM & 20 \\
stablediffusion1.4 & PNDM & 10 &
stablediffusion1.4-8bit & DDIM & 10 \\
stablediffusion1.4 & DDIM & 10  &
stablediffusion1.4-8bit & DDIM & 20 \\
stablediffusion1.4 & DDIM & 50 &
stablediffusion1.4-8bit & DDIM & 50 \\
stablediffusion1.4 & DPM & 10 &
stablediffusion1.4-8bit & DPM & 20 \\
stablediffusion1.4-6bit & DDIM & 10 &
stablediffusion1.5 & PNDM & 20 \\
stablediffusion1.4-6bit & DDIM & 50 &
stablediffusion1.5 & PNDM & 50 \\
stablediffusion1.4-6bit & DPM & 10 &
stablediffusion1.5 & DDIM & 50 \\
stablediffusion1.4-8bit & DPM & 10 &
stablediffusion1.5 & DPM & 10 \\
stablediffusion1.5 & PNDM & 10 &
stablediffusion1.5-6bit & DDIM & 10 \\
stablediffusion1.5 & DDIM & 10 &
stablediffusion1.5-6bit & DDIM & 50 \\
stablediffusion1.5 & DDIM & 20 &
stablediffusion1.5-6bit & DPM & 20 \\
stablediffusion1.5 & DPM & 20 &
stablediffusion1.5-8bit & DDIM & 20 \\
stablediffusion1.5-6bit & DDIM & 20 &
stablediffusion1.5-8bit & DPM & 10 \\
stablediffusion1.5-6bit & DPM & 10  &
stablediffusion1.5-8bit & DPM & 20 \\
stablediffusion1.5-8bit & DDIM & 10 &
small-stablediffusion1.5 & PNDM & 20 \\
stablediffusion1.5-8bit & DDIM & 50 &
small-stablediffusion1.5 & PNDM & 50 \\
small-stablediffusion1.5 & PNDM & 10& small-stablediffusion1.5 & DPM & 20 \\
small-stablediffusion1.5 & DDIM & 10 & stablediffusion2.1 & PNDM & 10 \\
small-stablediffusion1.5 & DDIM & 20 & stablediffusion2.1 & PNDM & 20 \\ 
small-stablediffusion1.5 & DDIM & 50 & stablediffusion2.1 & DDIM & 10 \\
small-stablediffusion1.5 & DPM & 10 & stablediffusion2.1 & DDIM & 50 \\ 
stablediffusion2.1 & PNDM & 50 & stablediffusion2.1 & DPM & 10 \\ 
stablediffusion2.1 & DDIM & 20 & stablediffusion2.1 & DPM & 20 \\  
dreamlike-photoreal & PNDM & 20 & dreamlike-photoreal & PNDM & 10  \\ 
dreamlike-photoreal & DDIM & 10 & dreamlike-photoreal & PNDM & 50 \\
dreamlike-photoreal & DDIM & 20 & dreamlike-photoreal & DPM & 10 \\ 
dreamlike-photoreal & DDIM & 50 & dreamlike-photoreal & DPM & 20 \\
\midrule
\end{tabular}
\caption{\textbf{The detailed list of models in sub-task with "random split".} } 
\label{tab:random-split}
\end{table}

\begin{table}[h!]
\centering
\begin{tabular}{ccc|ccc}
\toprule
\multicolumn{3}{c|}{training models} & \multicolumn{3}{c}{testing models} \\ \cmidrule(l){1-6}
model & solver & step & model & solver & step\\ 
\midrule
\midrule
stablediffusion1.4 & PNDM & 10 & stablediffusion1.2 & PNDM & 10\\
stablediffusion1.4 & PNDM & 20 & stablediffusion1.2 & PNDM & 20\\
stablediffusion1.4 & PNDM & 50 & stablediffusion1.2 & PNDM & 50\\
stablediffusion1.4 & DDIM & 10 & stablediffusion1.2 & DDIM & 10\\
stablediffusion1.4 & DDIM & 20 & stablediffusion1.2 & DDIM & 20\\
stablediffusion1.4 & DDIM & 50 & stablediffusion1.2 & DDIM & 50\\
stablediffusion1.4 & DPM & 10 & stablediffusion1.2 & DPM & 10\\
stablediffusion1.4 & DPM & 20 & stablediffusion1.2 & DPM & 20\\
stablediffusion1.5 & PNDM & 10 & stablediffusion1.2-6bit & DDIM & 10\\
stablediffusion1.5 & PNDM & 20 & stablediffusion1.2-6bit & DDIM & 20\\
stablediffusion1.5 & PNDM & 50 & stablediffusion1.2-6bit & DDIM & 50\\
stablediffusion1.5 & DDIM & 10 & stablediffusion1.2-6bit & DPM & 10\\
stablediffusion1.5 & DDIM & 20 & stablediffusion1.2-6bit & DPM & 20\\
stablediffusion1.5 & DDIM & 50 & stablediffusion1.2-8bit & DDIM & 10\\
stablediffusion1.5 & DPM & 10 & stablediffusion1.2-8bit & DDIM & 20\\
stablediffusion1.5 & DPM & 20 & stablediffusion1.2-8bit & DDIM & 50\\
stablediffusion1.5-6bit & DDIM & 10 & stablediffusion1.2-8bit & DPM & 10\\
stablediffusion1.5-6bit & DDIM & 20 & stablediffusion1.2-8bit & DPM & 20\\
stablediffusion1.5-6bit & DDIM & 50 & stablediffusion1.4-6bit & DDIM & 10\\
stablediffusion1.5-6bit & DPM & 10 & stablediffusion1.4-6bit & DDIM & 20\\
stablediffusion1.5-6bit & DPM & 20 & stablediffusion1.4-6bit & DDIM & 50\\
stablediffusion1.5-8bit & DDIM & 10 & stablediffusion1.4-6bit & DPM & 10\\
stablediffusion1.5-8bit & DDIM & 20 & stablediffusion1.4-6bit & DPM & 20\\
stablediffusion1.5-8bit & DDIM & 50 & stablediffusion1.4-8bit & DDIM & 10\\
stablediffusion1.5-8bit & DPM & 10 & stablediffusion1.4-8bit & DDIM & 20\\
stablediffusion1.5-8bit & DPM & 20 & stablediffusion1.4-8bit & DDIM & 50\\
stablediffusion2.1 & PNDM & 10 & stablediffusion1.4-8bit & DPM & 10\\
stablediffusion2.1 & PNDM & 20 & stablediffusion1.4-8bit & DPM & 20\\
stablediffusion2.1 & PNDM & 50 & small-stablediffusion1.5 & PNDM & 10\\
stablediffusion2.1 & DDIM & 10 & small-stablediffusion1.5 & PNDM & 20\\
stablediffusion2.1 & DDIM & 20 & small-stablediffusion1.5 & PNDM & 50\\
stablediffusion2.1 & DDIM & 50 & small-stablediffusion1.5 & DDIM & 10\\
stablediffusion2.1 & DPM & 10 & small-stablediffusion1.5 & DDIM & 20\\
stablediffusion2.1 & DPM & 20 & small-stablediffusion1.5 & DDIM & 50\\
 &  &  & small-stablediffusion1.5 & DPM & 10\\
 &  &  & small-stablediffusion1.5 & DPM & 20\\
 &  &  & dreamlike-photoreal & PNDM & 10\\
 &  &  & dreamlike-photoreal & PNDM & 20\\
 &  &  & dreamlike-photoreal & PNDM & 50\\
 &  &  & dreamlike-photoreal & DDIM & 10\\
 &  &  & dreamlike-photoreal & DDIM & 20\\
 &  &  & dreamlike-photoreal & DDIM & 50\\
 &  &  & dreamlike-photoreal & DPM & 10\\
 &  &  & dreamlike-photoreal & DPM & 20\\
\midrule
\end{tabular}
\caption{\textbf{The detailed list of models in sub-task with "across model variances".} } 
\label{tab:across-model}
\end{table}

\begin{table}[h!]
\centering
\begin{tabular}{ccc|ccc}
\toprule
\multicolumn{3}{c|}{training models} & \multicolumn{3}{c}{testing models} \\ \cmidrule(l){1-6}
model & solver & step & model & solver & step\\ 
\midrule
\midrule
stablediffusion1.2 & PNDM & 10 & stablediffusion1.2 & PNDM & 20\\
stablediffusion1.2 & DDIM & 10 & stablediffusion1.2 & PNDM & 50\\
stablediffusion1.2 & DDIM & 20 & stablediffusion1.2 & DPM & 10\\
stablediffusion1.2 & DDIM & 50 & stablediffusion1.2 & DPM & 20\\
stablediffusion1.2-6bit & DDIM & 10 & stablediffusion1.2-6bit & DDIM & 50\\
stablediffusion1.2-6bit & DDIM & 20 & stablediffusion1.2-6bit & DPM & 20\\
stablediffusion1.2-6bit & DPM & 10 & stablediffusion1.2-8bit & DDIM & 50\\
stablediffusion1.2-8bit & DDIM & 10 & stablediffusion1.2-8bit & DPM & 20\\
stablediffusion1.2-8bit & DDIM & 20 & stablediffusion1.4 & PNDM & 20\\
stablediffusion1.2-8bit & DPM & 10 & stablediffusion1.4 & PNDM & 50\\
stablediffusion1.4 & PNDM & 10 & stablediffusion1.4 & DPM & 10\\
stablediffusion1.4 & DDIM & 10 & stablediffusion1.4 & DPM & 20\\
stablediffusion1.4 & DDIM & 20 & stablediffusion1.4-6bit & DDIM & 50\\
stablediffusion1.4 & DDIM & 50 & stablediffusion1.4-6bit & DPM & 20\\
stablediffusion1.4-6bit & DDIM & 10 & stablediffusion1.4-8bit & DDIM & 50\\
stablediffusion1.4-6bit & DDIM & 20 & stablediffusion1.4-8bit & DPM & 20\\
stablediffusion1.4-6bit & DPM & 10 & stablediffusion1.5 & PNDM & 20\\
stablediffusion1.4-8bit & DDIM & 10 & stablediffusion1.5 & PNDM & 50\\
stablediffusion1.4-8bit & DDIM & 20 & stablediffusion1.5 & DPM & 10\\
stablediffusion1.4-8bit & DPM & 10 & stablediffusion1.5 & DPM & 20\\
stablediffusion1.5 & PNDM & 10 & stablediffusion1.5-6bit & DDIM & 50\\
stablediffusion1.5 & DDIM & 10 & stablediffusion1.5-6bit & DPM & 20\\
stablediffusion1.5 & DDIM & 20 & stablediffusion1.5-8bit & DDIM & 50\\
stablediffusion1.5 & DDIM & 50 & stablediffusion1.5-8bit & DPM & 20\\
stablediffusion1.5-6bit & DDIM & 10 & small-stablediffusion1.5 & PNDM & 20\\
stablediffusion1.5-6bit & DDIM & 20 & small-stablediffusion1.5 & PNDM & 50\\
stablediffusion1.5-6bit & DPM & 10 & small-stablediffusion1.5 & DPM & 10\\
stablediffusion1.5-8bit & DDIM & 10 & small-stablediffusion1.5 & DPM & 20\\
stablediffusion1.5-8bit & DDIM & 20 & stablediffusion2.1 & PNDM & 20\\
stablediffusion1.5-8bit & DPM & 10 & stablediffusion2.1 & PNDM & 50\\
small-stablediffusion1.5 & PNDM & 10 & stablediffusion2.1 & DPM & 10\\
small-stablediffusion1.5 & DDIM & 10 & stablediffusion2.1 & DPM & 20\\
small-stablediffusion1.5 & DDIM & 20 & dreamlike-photoreal & PNDM & 20\\
small-stablediffusion1.5 & DDIM & 50 & dreamlike-photoreal & PNDM & 50\\
stablediffusion2.1 & PNDM & 10 & dreamlike-photoreal & DPM & 10\\
stablediffusion2.1 & DDIM & 10 & dreamlike-photoreal & DPM & 20\\
stablediffusion2.1 & DDIM & 20 & \\
stablediffusion2.1 & DDIM & 50 & \\
dreamlike-photoreal & PNDM & 10 & \\
dreamlike-photoreal & DDIM & 10 & \\
dreamlike-photoreal & DDIM & 20 & \\
dreamlike-photoreal & DDIM & 50 & \\
\midrule
\end{tabular}
\caption{\textbf{The detailed list of models in sub-task with "across schedules".} } 
\label{tab:across-schedules}
\end{table}

\subsection{Results under Different Model Settings Split on COCO annotations}
\begin{figure*}[h!]
    \centering
    \includegraphics[width=0.9\linewidth]{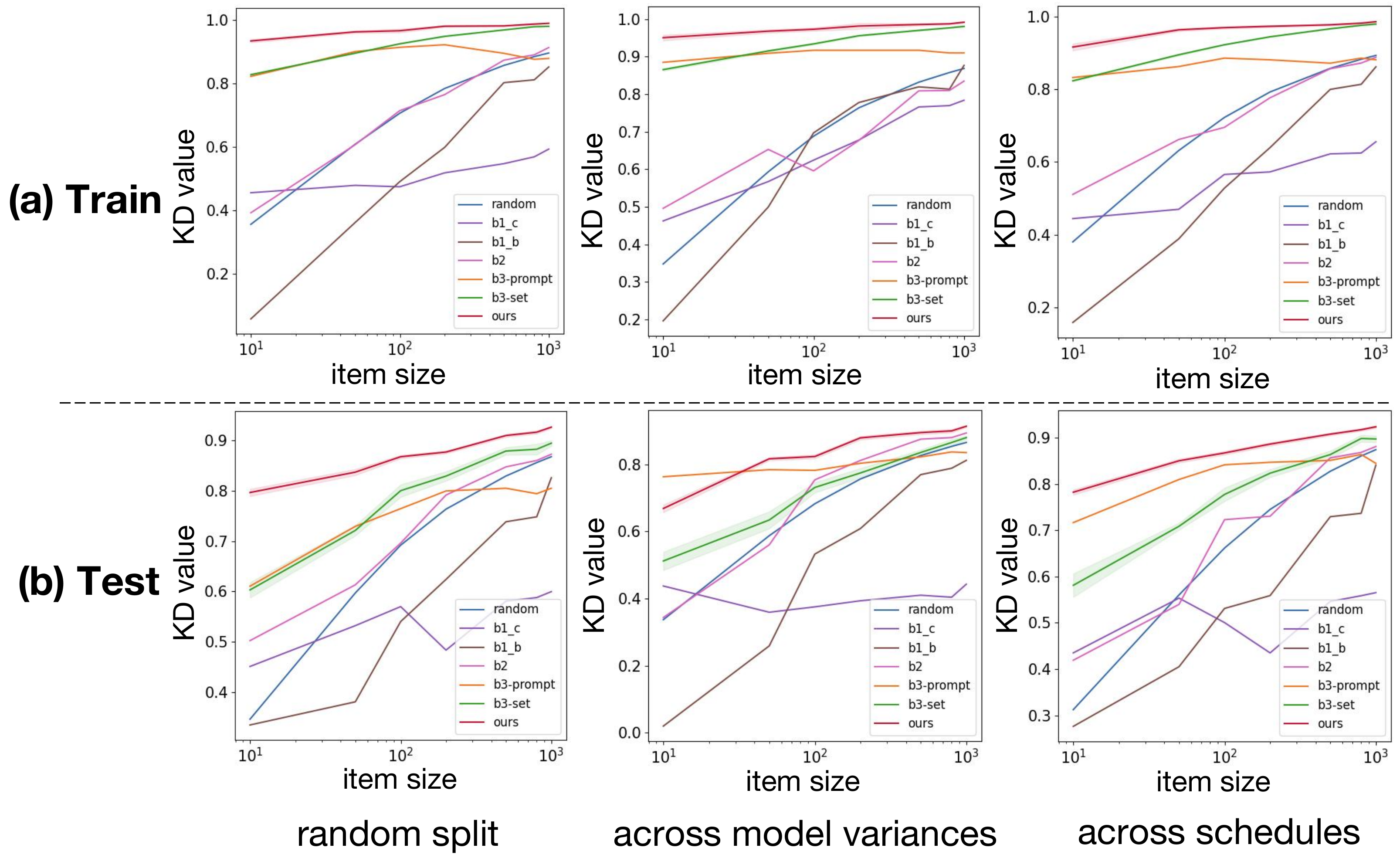}
    \caption{\textbf{Comparisons with all baselines of the Kendall’s Tau for CLIP-Score on COCO dataset.}}
    \label{fig:CLIP_coco_com}
\end{figure*}

\begin{figure*}[h!]
    \centering
    \includegraphics[width=0.9\linewidth]{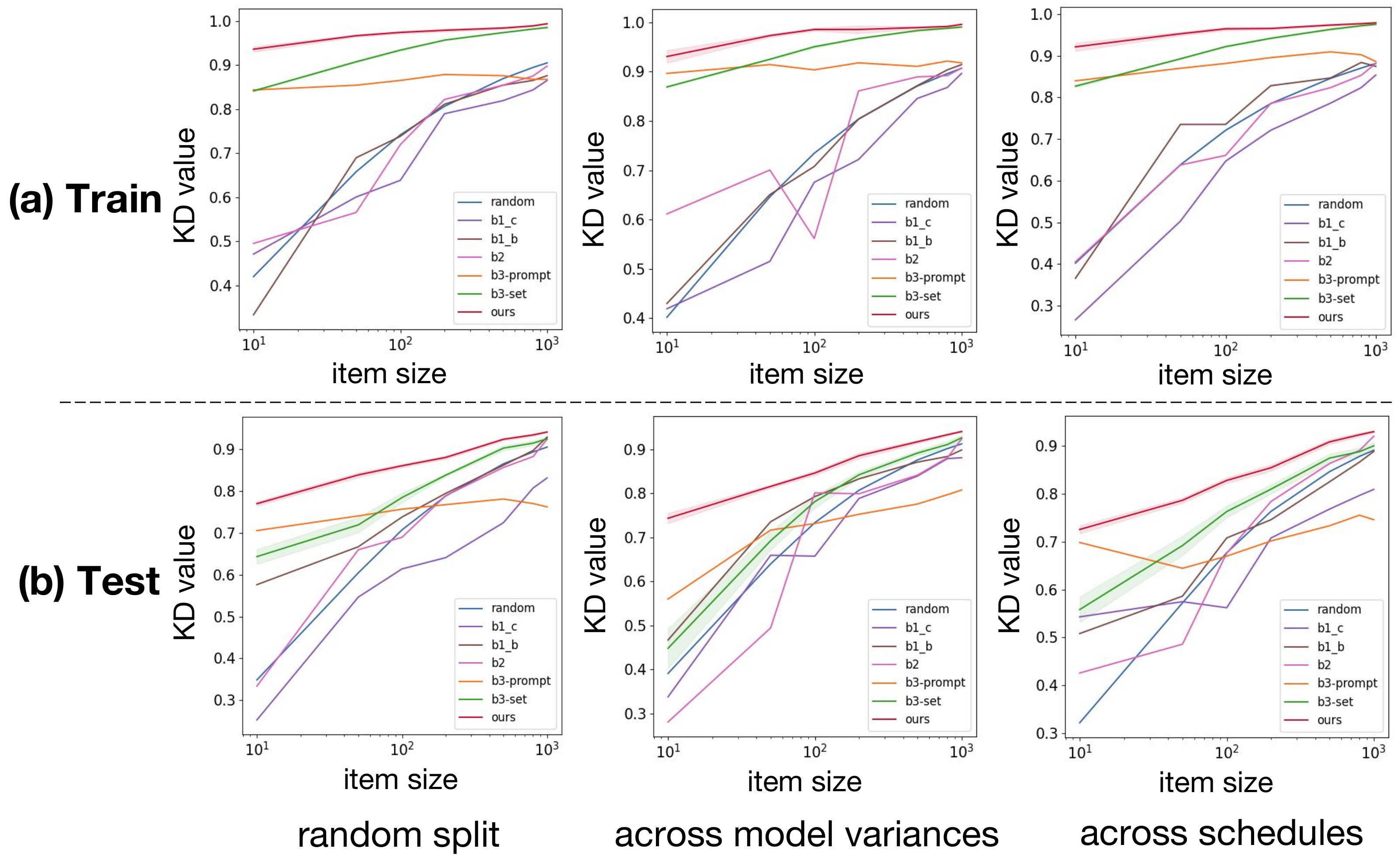}
    \caption{\textbf{Comparisons with all baselines of the Kendall’s Tau for ImageReward on COCO dataset.}}
    \label{fig:ImageReward_coco_com}
\end{figure*}

\begin{figure*}[h!]
    \centering
    \includegraphics[width=0.9\linewidth]{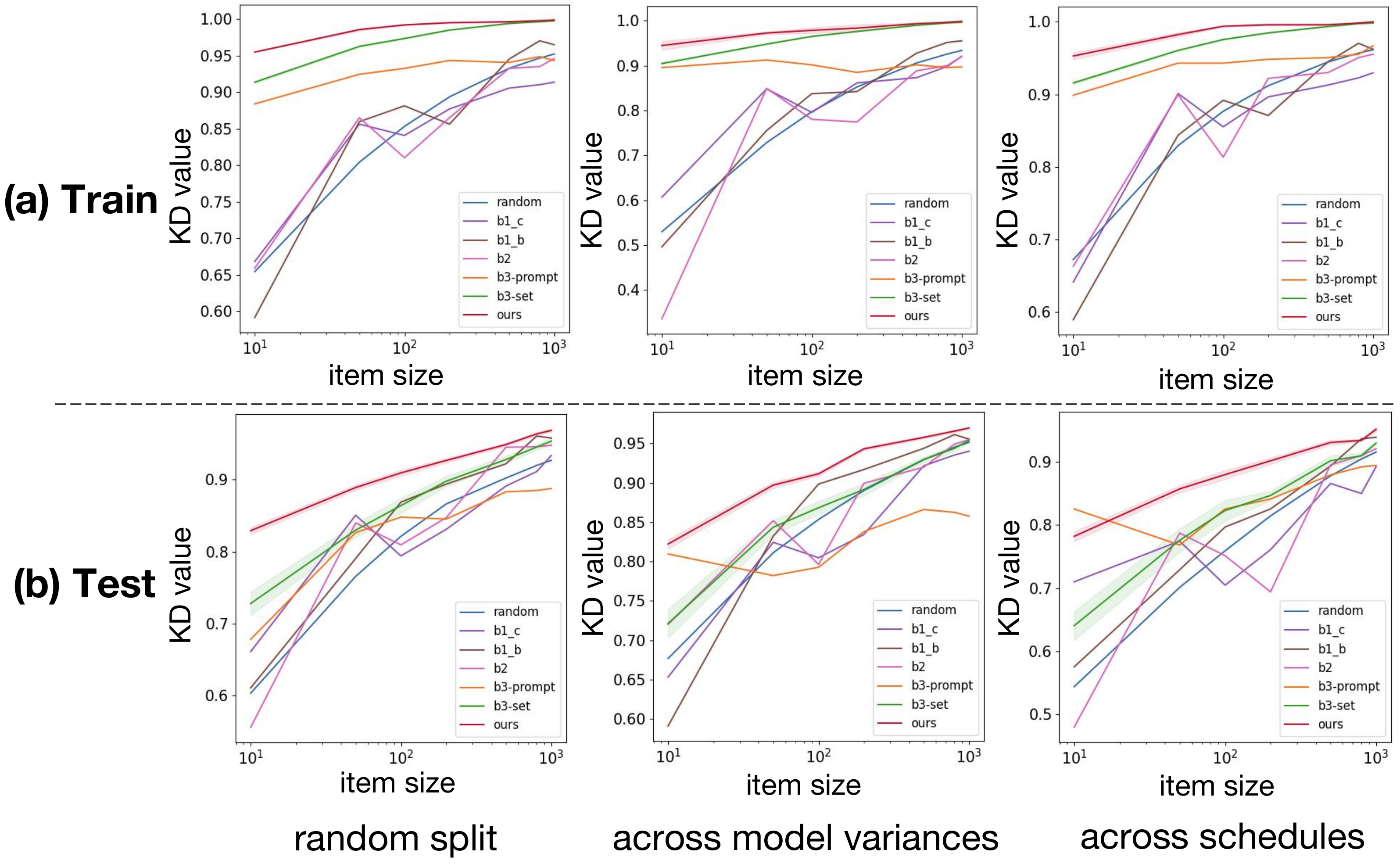}
    \caption{\textbf{Comparisons with all baselines of the Kendall’s Tau for HPS on COCO dataset.}}
    \label{fig:HPS_coco_com}
\end{figure*}

\begin{figure*}[h!]
    \centering
    \includegraphics[width=0.9\linewidth]{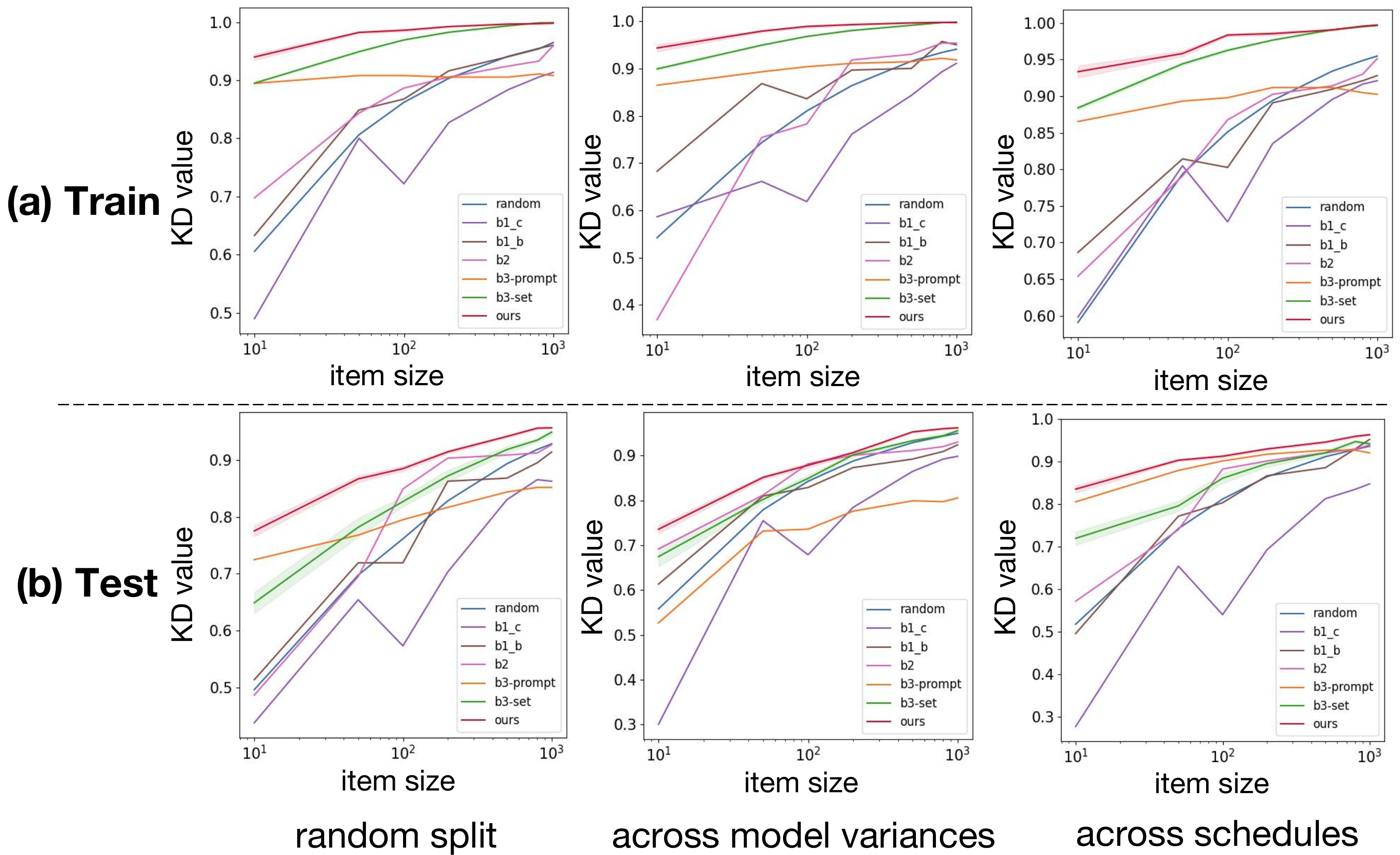}
    \caption{\textbf{Comparisons with all baselines of the Kendall’s Tau for Aesthetic on COCO dataset.}}
    \label{fig:Aesthetic_coco_com}
\end{figure*}

\begin{figure*}[h!]
    \centering
    \includegraphics[width=0.9\linewidth]{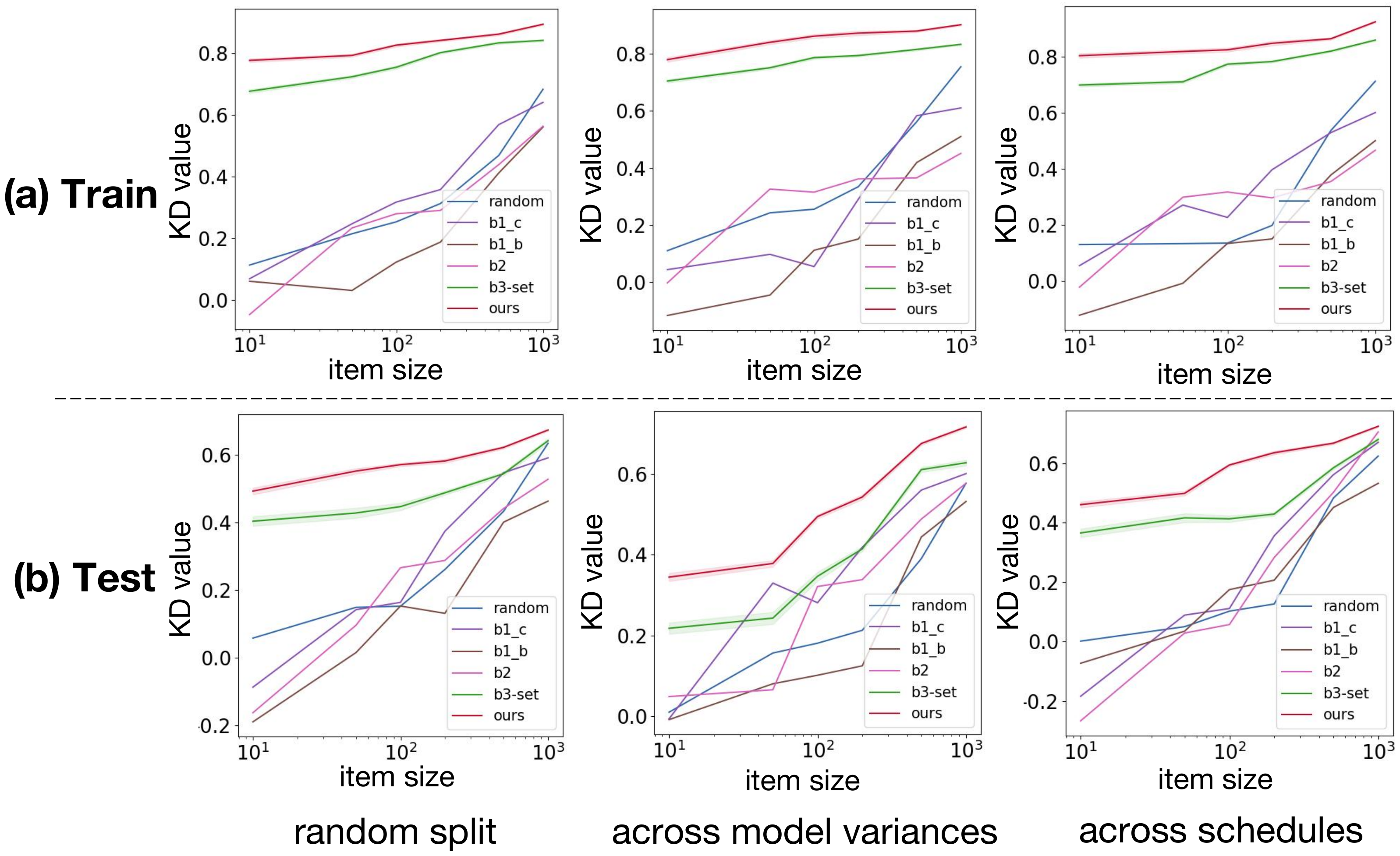}
    \caption{\textbf{Comparisons with all baselines of the Kendall’s Tau for FID on COCO dataset.}}
    \label{fig:FID_coco_com}
\end{figure*}

We present the comparative results of all methods across various metrics in \cref{fig:CLIP_coco_com,fig:ImageReward_coco_com,fig:HPS_coco_com,fig:Aesthetic_coco_com,fig:FID_coco_com}. It is evident that \method consistently outperforms others in all scenarios, which demonstrates the generalization ability of FlashEval across diverse model or schedules.

\subsection{Results under Different Model Settings Split on DiffusionDB }
As mentioned in Sec. 5.1 of the main paper, due to the substantial size of the complete diffusionDB dataset, it is not feasible to iterate through all prompts (2,000,000) to get the ground-truth performance. Considering the cost, we select a relatively large amount of samples (5000) and use the averaged performance on them as the proxy ``ground-truth'' of DiffusionDB (treating the 5,000 prompts as the ``whole set'' to further condense it into smaller representative subset). 

\begin{figure*}[h!]
    \centering
    \includegraphics[width=0.9\linewidth]{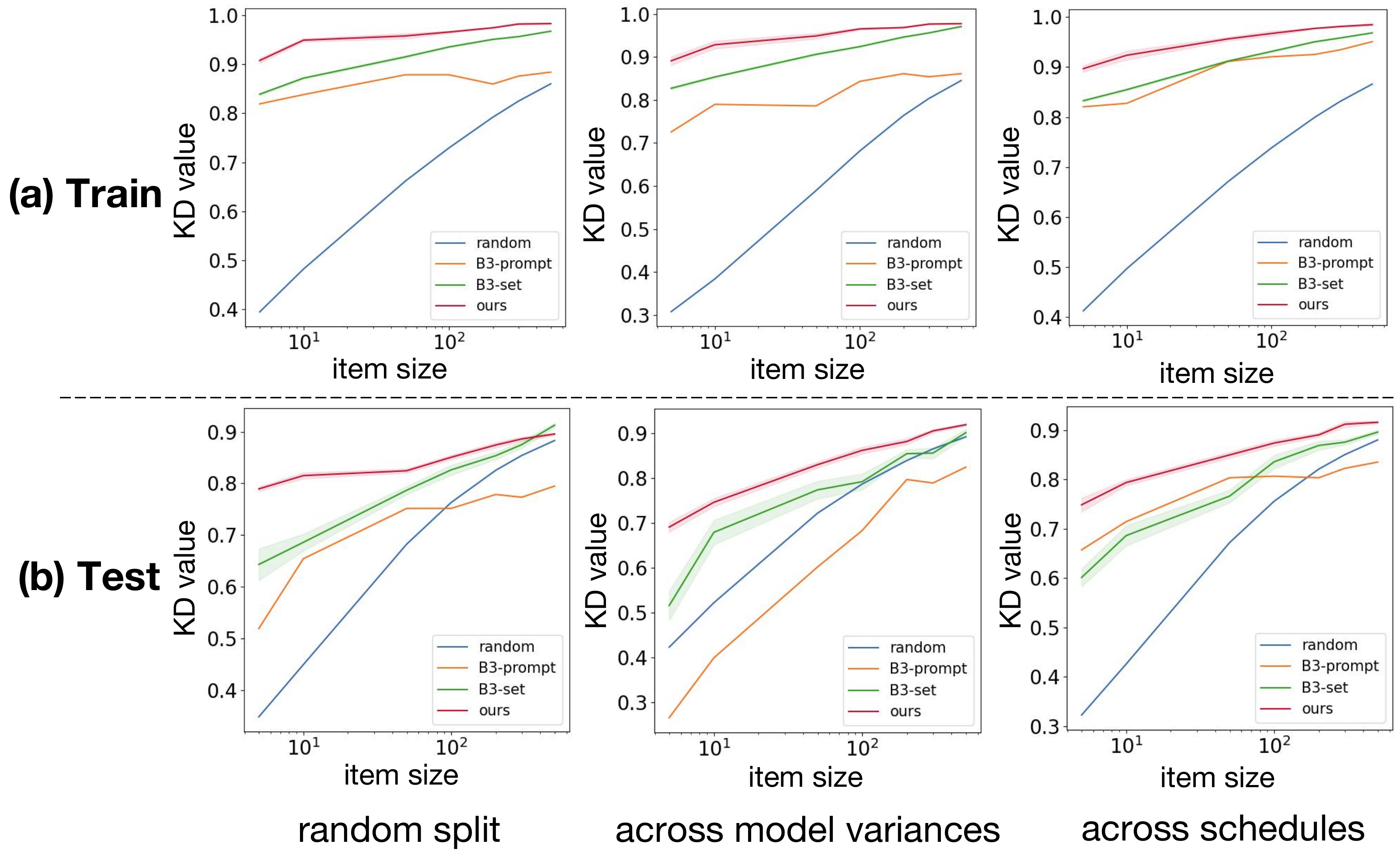}
    \caption{\textbf{Comparisons with metric space baselines of the Kendall’s Tau for CLIP-Score on DiffusionDB dataset.}}
    \label{fig:CLIP_diffusionDB_com}
\end{figure*}

\begin{figure*}[h!]
    \centering
    \includegraphics[width=0.9\linewidth]{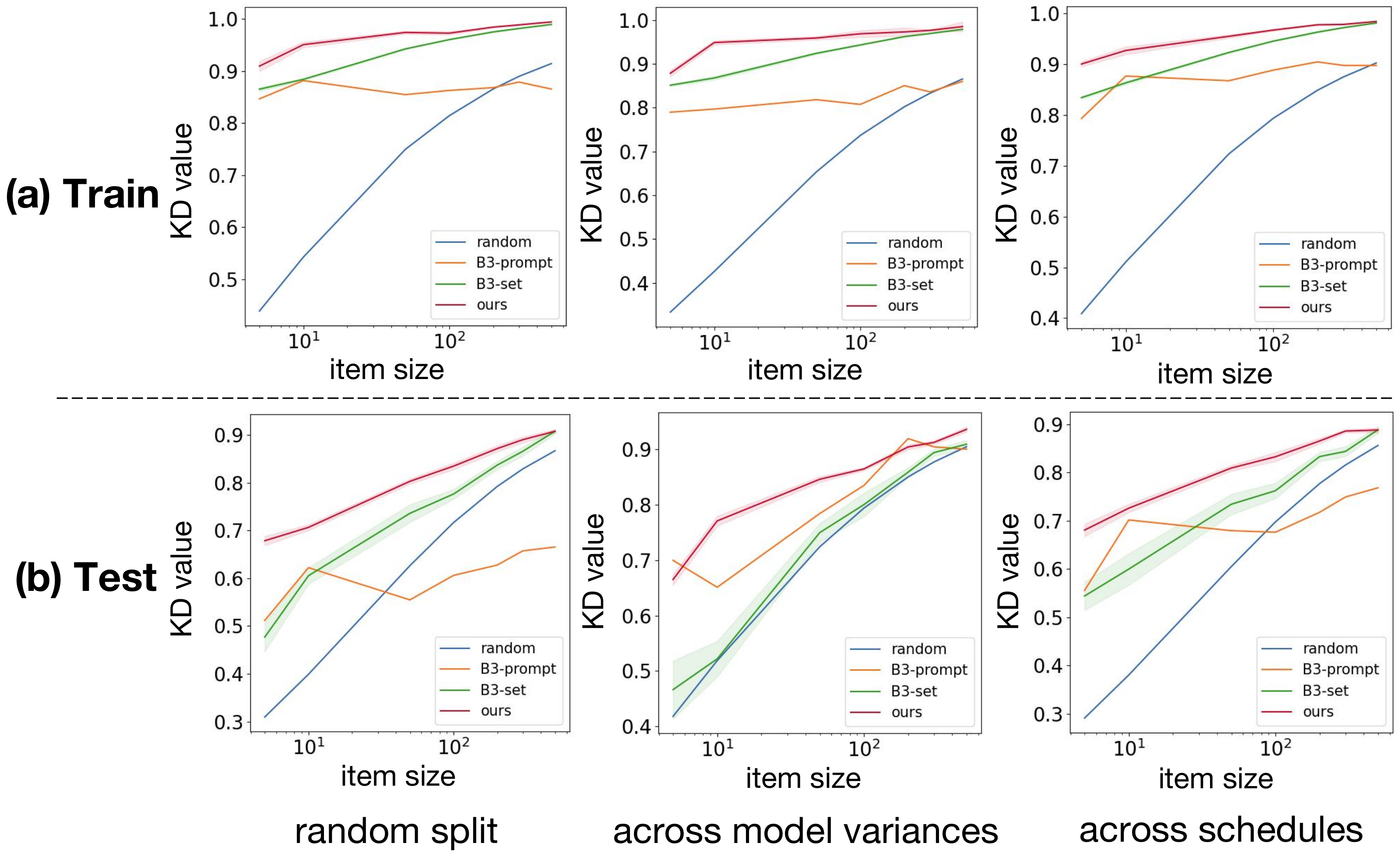}
    \caption{\textbf{Comparisons with metric space baselines of the Kendall’s Tau for ImageReward on DiffusionDB dataset.}}
    \label{fig:ImageReward_diffusionDB_com}
\end{figure*}

\begin{figure*}[h!]
    \centering
    \includegraphics[width=0.9\linewidth]{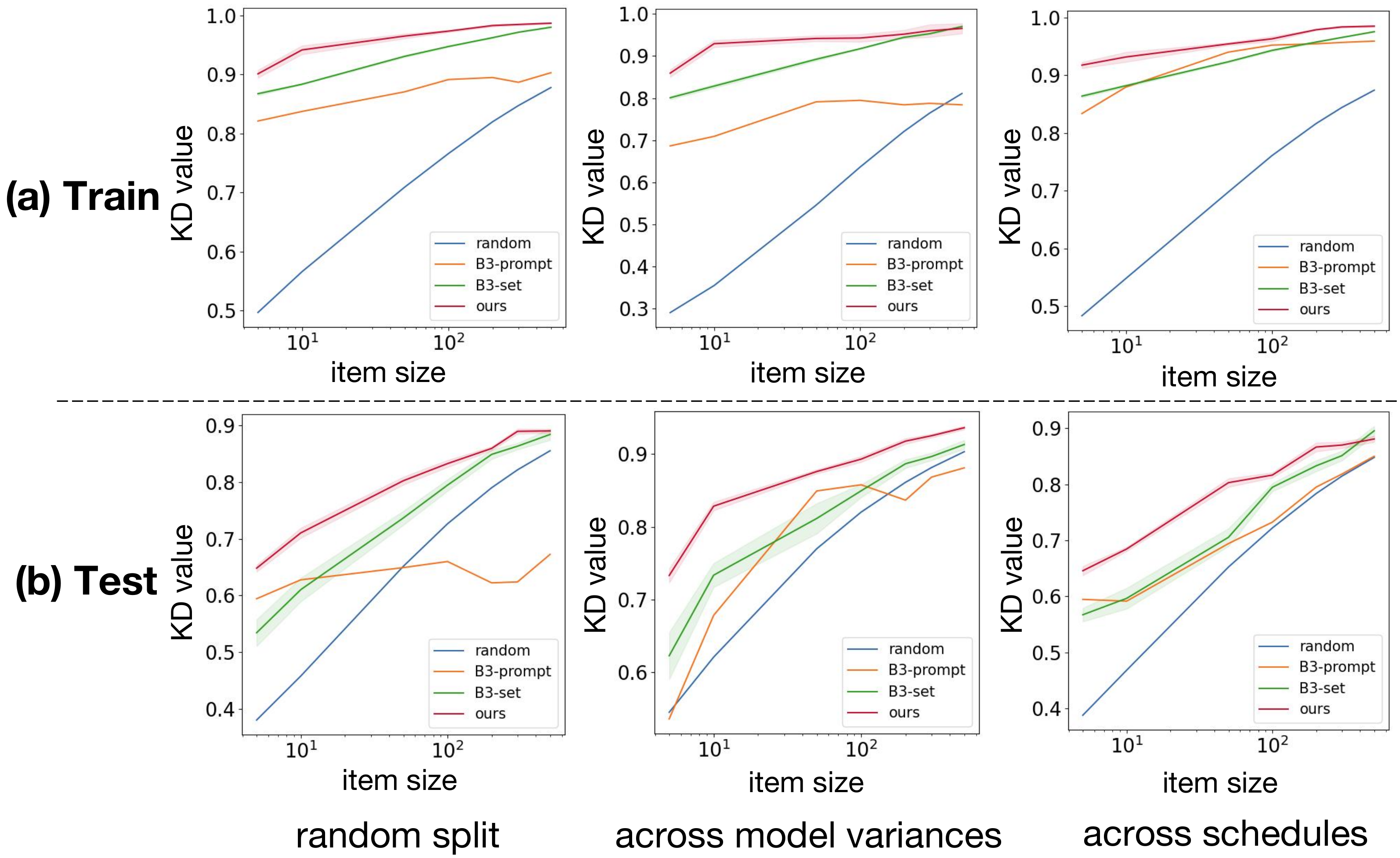}
    \caption{\textbf{Comparisons with metric space baselines of the Kendall’s Tau for HPS on DiffusionDB dataset.}}
    \label{fig:HPS_diffusionDB_com}
\end{figure*}

\begin{figure*}[h!]
    \centering
    \includegraphics[width=0.9\linewidth]{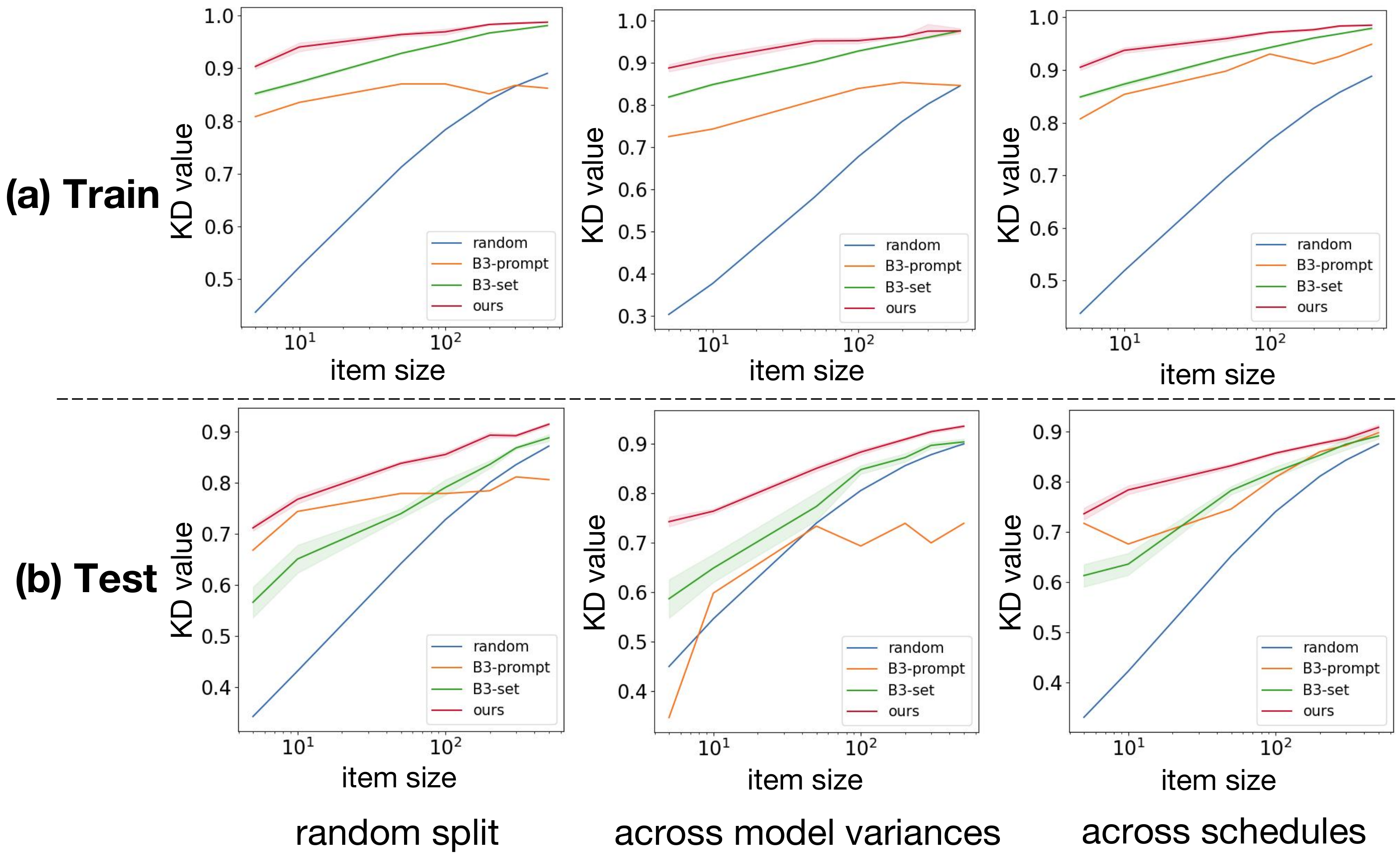}
    \caption{\textbf{Comparisons with metric space baselines of the Kendall’s Tau for Aesthetic on DiffusionDB dataset.}}
    \label{fig:Aesthetic_diffusionDB_com}
\end{figure*}

\begin{figure*}[h!]
    \centering
    \includegraphics[width=0.9\linewidth]{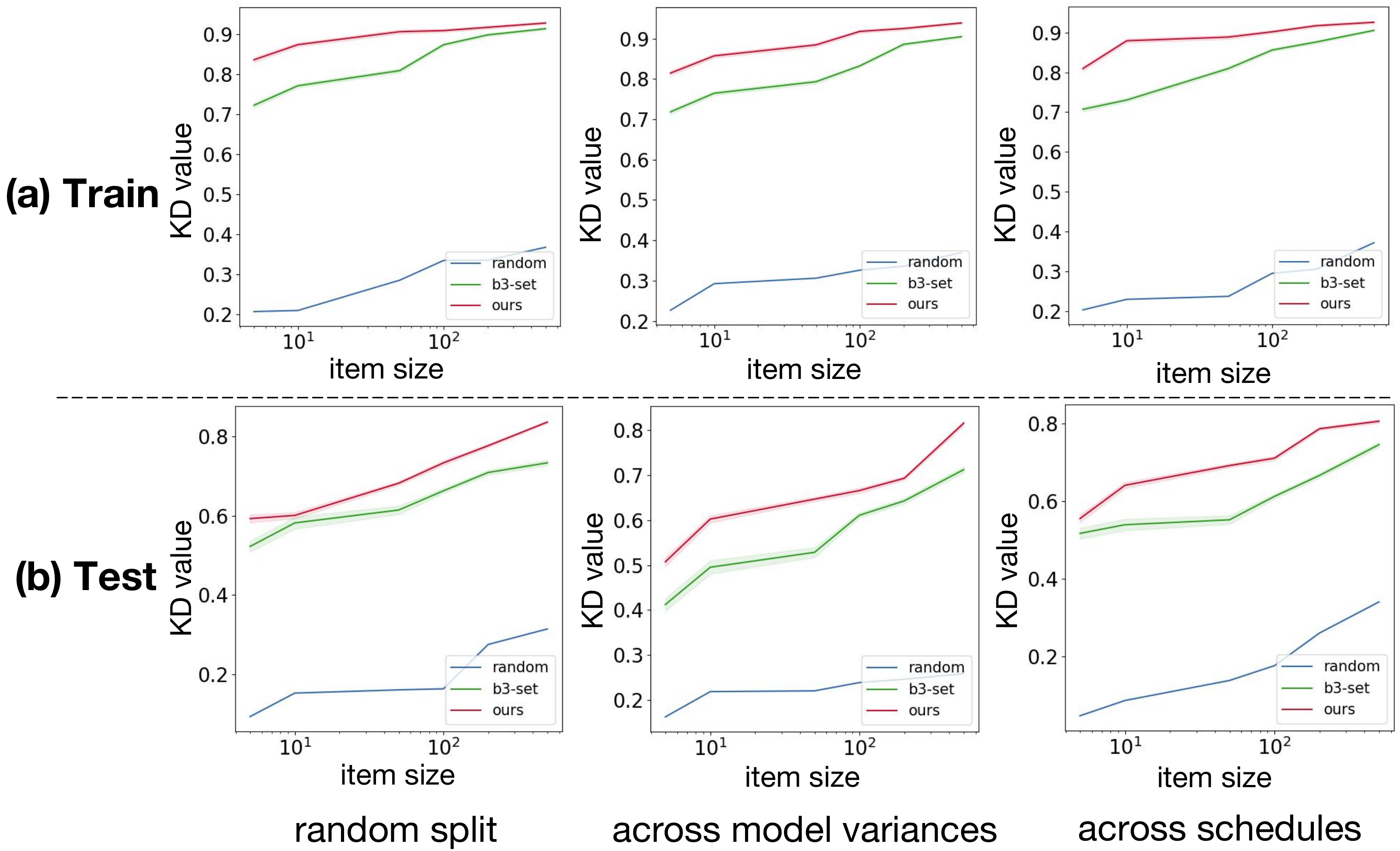}
    \caption{\textbf{Comparisons with metric space baselines of the Kendall’s Tau for FID on DiffusionDB dataset.}}
    \label{fig:FID_diffusionDB_com}
\end{figure*}

As could be witnessed from the \cref{fig:CLIP_diffusionDB_com,fig:HPS_diffusionDB_com,fig:ImageReward_diffusionDB_com,fig:Aesthetic_diffusionDB_com,fig:FID_diffusionDB_com}, our acquired subsets achieves superior evaluation compared with the random sample baseline (adopted by recent literature Lee \etal [41]) quality for different metrics across different model splits.

\clearpage

%
%

